\documentclass[nohyperref]{article}
\usepackage{subcaption}
\usepackage[table]{xcolor}

\usepackage{microtype}
\usepackage{graphicx}
\usepackage{booktabs} %

\usepackage{hyperref}

\usepackage[accepted]{icml2022}

\usepackage{amsmath}
\usepackage{amssymb}
\usepackage{mathtools}
\usepackage{amsthm}

\usepackage{xcolor}
\usepackage{color}
\usepackage{enumitem}
\usepackage{multirow}
\usepackage{makecell}

\usepackage[capitalize,noabbrev]{cleveref}

\usepackage{pifont}%
\newcommand{\cmark}{\ding{51}}%

\usepackage{amsmath}

\usepackage{xspace}
\newcommand*{\eg}{\emph{e.g.}\@\xspace}

\newcommand*{\model}{RPC}
\newcommand*{\aug}{WOLFMix}
\newcommand*{\dataset}{ModelNet-C}

\theoremstyle{plain}

\theoremstyle{definition}

\theoremstyle{remark}

\usepackage[textsize=tiny]{todonotes}

\icmltitlerunning{Benchmarking and Analyzing Point Cloud Classification under Corruptions}

\begin{document}

\twocolumn[
\icmltitle{Benchmarking and Analyzing Point Cloud Classification under Corruptions}

\icmlsetsymbol{equal}{*}

\begin{icmlauthorlist}
\icmlauthor{Jiawei Ren}{ntu}
\icmlauthor{Liang Pan}{ntu}
\icmlauthor{Ziwei Liu}{ntu}
\end{icmlauthorlist}

\icmlaffiliation{ntu}{S-Lab, Nanyang Technological University}

\icmlcorrespondingauthor{Ziwei Liu}{ziwei.liu@ntu.edu.sg}

\icmlkeywords{Machine Learning, ICML}

\vskip 0.3in
]

\printAffiliationsAndNotice{}  %

\begin{abstract}
3D perception, especially point cloud classification, has achieved substantial progress. 
However, in real-world deployment, point cloud corruptions are inevitable due to the scene complexity, sensor inaccuracy, and processing imprecision. In this work, we aim to rigorously benchmark and analyze point cloud classification under corruptions. To conduct a systematic investigation, we first provide a taxonomy of common 3D corruptions and identify the atomic corruptions. Then, we perform a comprehensive evaluation on a wide range of representative point cloud models to understand their robustness and generalizability. 
Our benchmark results show that although point cloud classification performance improves over time, the state-of-the-art methods are on the verge of being less robust.
Based on the obtained observations, we propose several effective techniques to enhance point cloud classifier robustness. We hope our comprehensive benchmark, in-depth analysis, and proposed techniques could spark future research in robust 3D perception.
Code is available at \url{https://github.com/jiawei-ren/modelnetc}.
\end{abstract}
\section{Introduction}

Robustness to common corruptions is crucial to point cloud classification. 
Compared to 2D images, point cloud data suffer more severe corruptions in real-world deployment due to the inaccuracy in 3D sensors and complexity in real-world 3D scenes~\cite{wu2019squeezesegv2, yan2020pointasnl}. 
Furthermore, point cloud is widely employed in safety-critical applications such as autonomous driving. 
Therefore, robustness to out-of-distribution (OOD) point cloud data caused by corruptions becomes an important part of the test suite since the beginning of learning-based point cloud classification~\cite{qi2017pointnet, simonovsky2017ecc}.

\begin{figure}[t!]
\vskip 0.in
\begin{center}
\centerline{\includegraphics[width=\columnwidth]{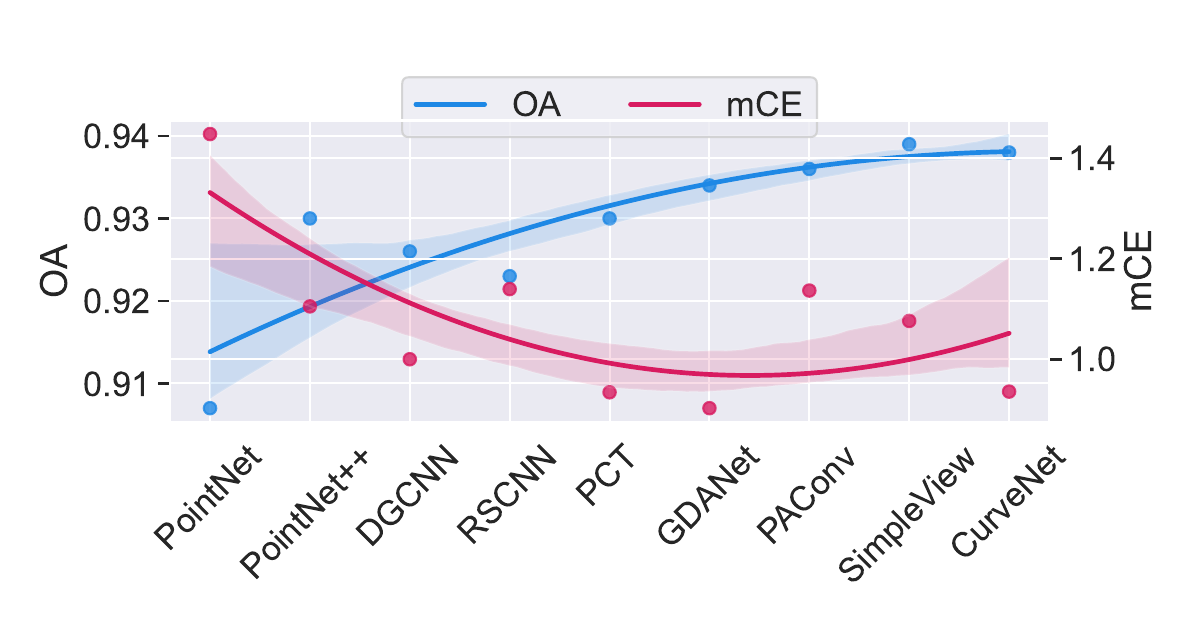}}
\vskip -0.4in
\end{center}
\vskip -0.0in
\end{figure}
\begin{figure}[t!]
\vskip 0.in
\begin{center}
\centerline{\includegraphics[width=\columnwidth]{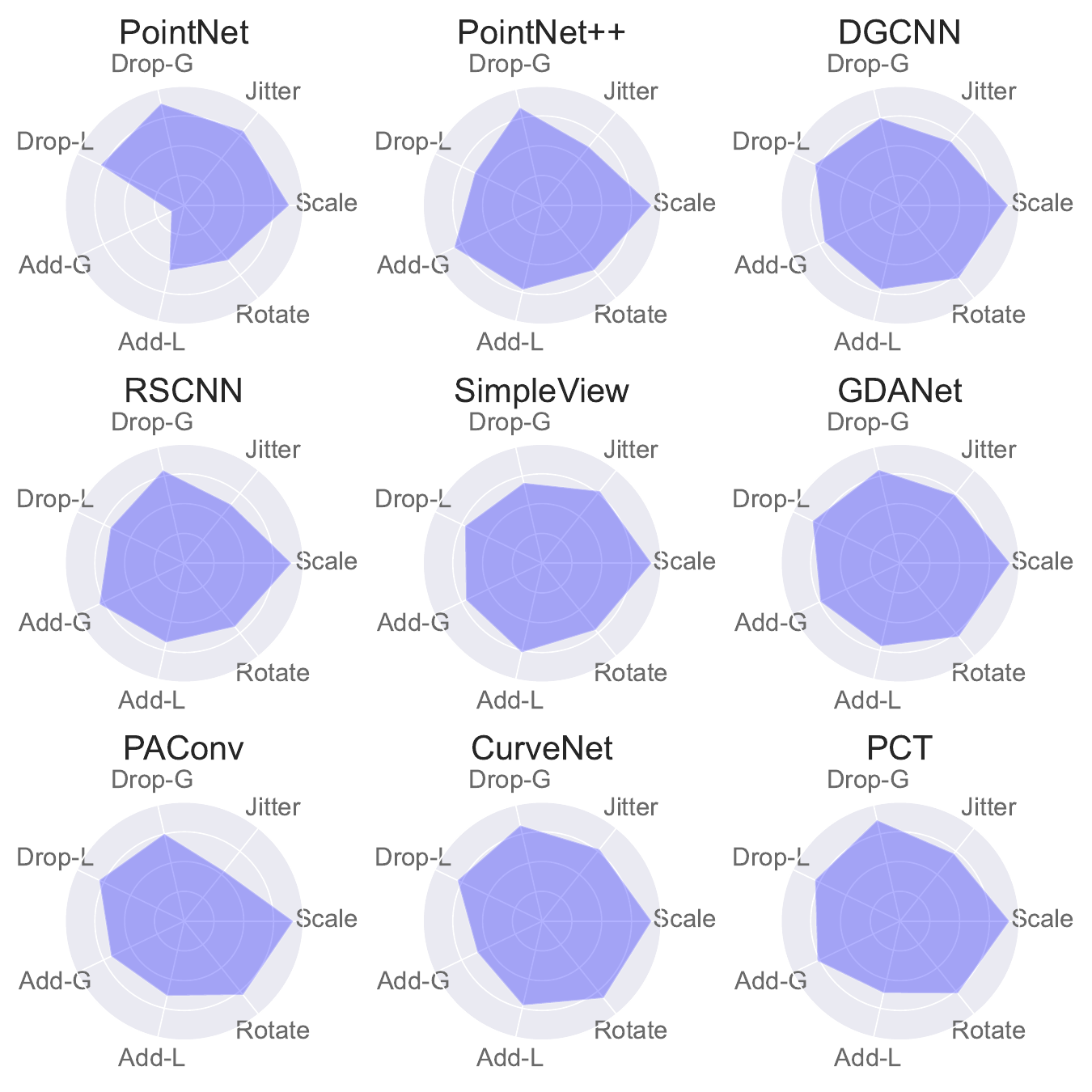}}
\vspace{-5mm}
\caption{\textbf{Upper.} Blue curve shows overall accuracy (OA) on ModelNet40. Red curve shows mean Corruption Error (mCE) on proposed \dataset{}. Methods are sorted in chronological order. OA gradually saturates but mCE is at the risk of increasing due to the lack of a standard test suite. \textbf{Lower.} Point cloud classifer's robustness to various corruptions in a radar chart. Proposed \dataset{} allows fine-grained corruption analysis. Different architectures have diverse strengths and weaknesses to corruptions. "-G": -Global. "-L": -Local.}
\label{fig:radar}
\end{center}
\vskip -0.3in
\end{figure}

\begin{table*}[t!]
\setlength{\tabcolsep}{8.5pt}
\centering\scriptsize 
\caption{Corruptions studied in existing robustness analysis. Prior works evaluate point cloud classification robustness on different sets of corruptions, and hence their evaluations can be partial and unfair. To standardize the corruption evaluation, our test suite \emph{\dataset{}} includes all previously studied corruptions, including ``Jitter'', ``Drop Global/Local'', ``Add Global/Local'', ``Scale'' and ``Rotate''.}
\label{tab:prior_robust}
\begin{tabular}{lccccccc}
\toprule
{} & \small Jitter & \small Drop Global & \small Drop Local & \small Add Global & \small Add Local & \small Scale & \small Rotate \\
\midrule
PointNet~\cite{qi2017pointnet}& \cmark & \cmark & {} & \cmark & {} & {} & {} \\
ECC~\cite{simonovsky2017ecc}& \cmark & \cmark & {} & {} & {} & {} & {} \\
PointNet++~\cite{qi2017pointnetplusplus} & {} & \cmark & {} & {} & {} & {} & {}\\
DGCNN~\cite{wang2019dgcnn} & {} & \cmark & {} & {} & {} & {} & {} \\
RSCNN~\cite{liu2019rscnn} & {} & \cmark & {} & {} & {} & {}  & \cmark \\
PointASNL~\cite{yan2020pointasnl} & {} & \cmark & {} & \cmark & {} & {} & {} \\
Orderly Disorder~\cite{ghahremani2020orderly} & \cmark & {} & {} & {} & {} & {} & {} \\
PointAugment~\cite{li2020pointaugment} & \cmark & \cmark & {} & {} & {} & \cmark & \cmark \\
PointMixup~\cite{chen2020pointmixup} & \cmark & \cmark & {} & {} & {} & \cmark & \cmark  \\
PAConv~\cite{xu2021paconv} & \cmark  & {} & {} & {} & {} & \cmark  & \cmark   \\
OcCo~\cite{wang2021occo} & {} & \cmark & {} & {} & {} & {} & {}\\
Triangle-Net~\cite{xiao2021triangle}  & \cmark & \cmark & {} & {} & {} & \cmark & \cmark \\
Curve-Net~\cite{xiang2021curvenet}  & \cmark & \cmark & {} & {} & {} & {} & {} \\
RSMix~\cite{lee2021rsmix}  & \cmark & \cmark & {} & {} & {} & \cmark & \cmark \\
PointWolf~\cite{kim2021pointwolf} & \cmark & \cmark & \cmark & {} & \cmark & {} & {}  \\
GDANet~\cite{xu2021gdanet} & {} & \cmark & {} & {} & {} & {} & \cmark  \\
\midrule
\emph{Our Benchmark (\textbf{\dataset{}})} & \cmark & \cmark & \cmark & \cmark & \cmark & \cmark & \cmark \\

\bottomrule
\end{tabular}
\end{table*}

Ideally, robustness should be measured in a standard way like how classification accuracy and computational cost are measured. However, prior research evaluates point cloud classifier robustness in many different protocols:

    \noindent\emph{Protocol-1.}
    Evaluate the robustness to a selected set of corruptions~\cite{qi2017pointnet, qi2017pointnetplusplus, wang2019dgcnn, chen2020pointmixup, kim2021pointwolf}, \eg, random point dropping and random jittering. This evaluation method is popular in point cloud research, as summarized in \autoref{tab:prior_robust}. However, the freedom to select corruptions brings both positive and negative effects to the evaluation. On the upside, customized selection allows the evaluation to focus on the most characteristic corruptions. 
    On the downside, a selected set of corruptions cannot provide a comprehensive evaluation of a model's robustness. In addition, different corruption selections and training protocols in implementation also make it difficult to compare across methods.
    
    \noindent\emph{Protocol-2.}
    Evaluate the robustness to the sim-to-real gap~\cite{reizenstein21co3d, ahmadyan2021objectron}, \eg, train on ModelNet40~\cite{wu20153d} and test on ScanObjectNN~\cite{uy2019scanobjectnn}. To exploit the naturally occurred corruptions in real-world point cloud object datasets, robustness is formulated as the generalizability from a synthetic training set to a real test set. However, real-world corruptions always come in a composite way, \eg, self-occlusion and scanner noise, making it hard to analyze each corruption independently. Besides, the sim-to-real performance gap couples with the domain gap within each category, \eg, a \emph{chair} in ModelNet40 and ScanObjectNN may have different styles, which obfuscates the evaluation results.
    
    \noindent\emph{Protocol-3.}
    Evaluate the robustness to adversarial attack~\cite{zhou2019dup, dong2020self, sun2021adversarially}, \eg, adversarial point shifting and dropping. Different from real-world scenarios where corruptions are drawn from natural distributions, adversarial attacks corrupt point clouds for the purpose to deceive a classifier while keeping the attacked point cloud similar to the input. Therefore, adversarial robustness is a good measure of a model's worst-case performance but can not reflect a point cloud classifier's robustness to common corruptions in the natural world.

\begin{figure*}[t!]
\vskip 0.in
\center
\begin{center}
\centerline{\includegraphics[width=1.8\columnwidth]{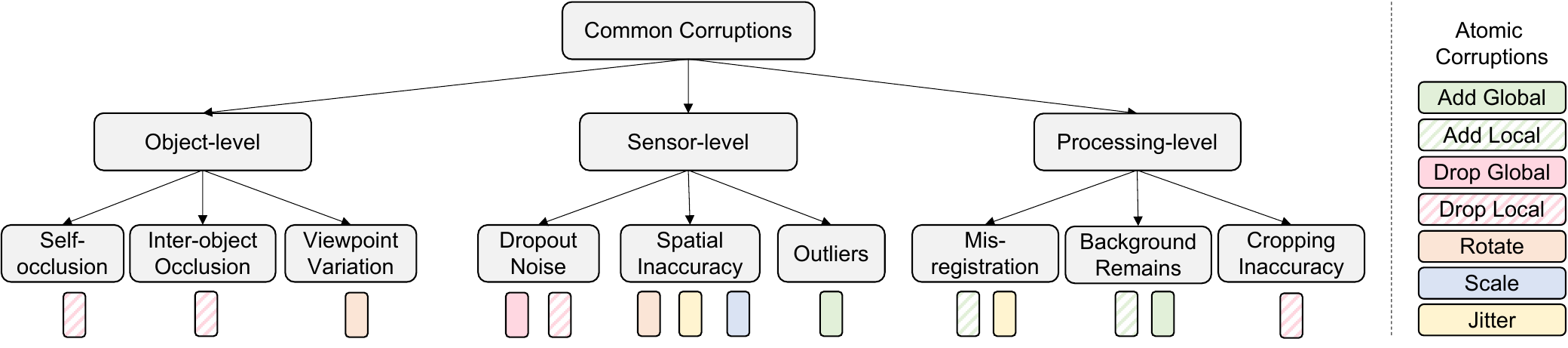}}
\vspace{-3mm}
\caption{Corruption taxonomy. We break down common corruptions into detailed corruption sources on object-, senor- and processing levels, which are further simplified into a combination of seven atomic corruptions for a more controllable empirical analysis.}
\label{fig:taxanomy}
\end{center}
\vskip -0.3in
\end{figure*}

Despite various ways to evaluate a point cloud classifier's robustness, there lacks a standard, comprehensive benchmark for point cloud classification under corruptions. In this work, we present a full corruption test suite to close this gap. First, we break down real-world corruptions in \emph{Protocol-2} into 7 fundamental atomic corruptions (~\autoref{fig:taxanomy}), which also forms a superset of the ad-hoc corruption selections in \emph{Protocol-1}. As we aim to measure real-world robustness, adversarial attacks in \emph{Protocol-3} are excluded. Then, we apply the atomic corruptions to the validation set of ModelNet40 as our corruption test suite dubbed \emph{\dataset{}}. 
Inspired by the 2D image classification robustness benchmark~\cite{hendrycks2019bench}, we further create 5 severity levels for each atomic corruption and use the mean Corruption Error (mCE) metric for evaluation. 
Finally, based on the test suite, we benchmark 14 point cloud classification methods, including 9 architectures, 3 augmentations, and 2 pretrains. As shown in \autoref{fig:radar}, our benchmark results show that although point cloud classification performance on the clean ModelNet40 improves by time, \emph{state-of-the-art (SoTA) methods are on the verge of being less robust}. 

To remedy the issue, we conduct an in-depth analysis of the benchmark results and summarize two effective techniques to enhance point cloud classifier robustness.
    Strictly following the best design choice summarized from the benchmark results, we present Robust Point cloud Classifier (\model{}), a robust network architecture for point cloud classification, which achieves the least mCE on \emph{\dataset{}} benchmark, and comparable overall accuracy on the clean ModelNet40 with the SoTAs.
    In particular,
    we present \aug{}, a strong augmentation baseline that exploits both deformation-based augmentation and mix-based augmentation to provide a stronger regularization. Empirically, \aug{} achieves the best robustness results compared to existing augmentation techniques.
According to our experiments,
the performance gain by augmentations does not equally transfer to all model architectures. We identify the best combination from existing methods, 
and call for model design that fully exploits the augmentation power.

Our contributions are summarized as: \textbf{1)} We present the first systematically-designed test-suite \emph{\dataset{}} for point cloud classifier under corruptions. \textbf{2)} We comprehensively benchmark existing methods on their robustness to corruptions. \textbf{3)} We summarize several effective techniques, such as \model{} and \aug{}, to enhance point cloud classifier's robustness and identify that the synergy between architecture and augmentation should be considered in future research.
\section{Related Works}
\noindent\textbf{Point Cloud Classification.}
Point cloud classification serves as an fundamental task for 3D understanding from raw hardware inputs. Point cloud classifier has diverse architectural designs. There are MLP-based models~\cite{qi2017pointnet, qi2017pointnetplusplus}, convolution-based models~\cite{liu2019rscnn, xu2021paconv}, graph-based models~\cite{simonovsky2017ecc, wang2019dgcnn} and recently proposed transformer-based models~\cite{guo2020pct, zhao2021point, mazur2021cloudtransformers}. Besides, there is a rising discussion on point cloud augmentation, including mix-based augmentations~\cite{chen2020pointmixup, lee2021rsmix}, deformation-based augmentations~\cite{kim2021pointwolf} and auto-augmentations~\cite{li2020pointaugment}. Moreover, self-supervised pre-train has drawn much research attention recently. Pre-trains obtained from pre-text tasks like occlusion reconstruction~\cite{wang2021occo} and mask inpainting~\cite{yu2021pointbert} provide better classification performance than random initialization.

\noindent\textbf{Robustness in Point Cloud.}
Several attempts are made to improve point cloud classifier's robustness. Triangle-Net~\cite{xiao2021triangle} designs feature extraction that is invariant to positional, rotational, and scaling disturbances. Although Triangle-Net achieves exceptional robustness under extreme corruptions, its performance on clean data is not on par with SoTA. PointASNL~\cite{yan2020pointasnl} introduces adaptive sampling and local-nonlocal modules to improve robustness. However, PointASNL takes a fixed number of points in implementation. 
Other works improve a model's adversarial robustness by denoising and upsampling~\cite{zhou2019dup}, voting on subsampled point clouds~\cite{liu2021pointguard}, exploiting local feature's relative position~\cite{dong2020self} and self-supervision~\cite{sun2021adversarially}. 
RobustPointSet~\cite{taghanaki2020robustpointset} evaluates the robustness of point cloud classifiers under different corruptions, and shows that basic data augmentations poorly generalize to ``unseen'' corruptions. However, our work shows that more advanced augmentation techniques, \eg, mixing and local deformation, can substantially improve the robustness.

\noindent\textbf{Robustness Benchmarks in Image Classification.}
Comprehensive robustness benchmark has been built for 2D image classification recently. ImageNet-C~\cite{hendrycks2019bench} corrupts the ImageNet~\cite{imagenet_cvpr09}'s test set with simulated corruptions like compression loss and motion blur. ObjectNet~\cite{barbu2019objectnet} collects a test set with rich variations in rotation, background and viewpoint. ImageNetV2~\cite{recht2019do} re-collects a test set following ImageNet's protocal and evaluates the performance gap due to the natural distribution shift. Moreover, ImageNet-A~\cite{hendrycks2021nae} and  ImageNet-R~\cite{hendrycks2021many} benchmark classifier's robustness to natural adversarial examples and abstract visual renditions.

\section{Corruptions Taxonomy and Test Suite}
\subsection{Corruptions Taxonomy}
Real-world corruptions come from a wide range of sources, based on which we provide a taxonomy of the corruptions in \autoref{fig:taxanomy}. 
Common corruptions are categorized into three levels: object-level, sensor-level, and processing-level corruptions. 
Object-level corruptions come inherently in complex 3D scenes, where an object can be occluded by other objects or parts of itself. 
Different viewpoints also introduce variations to the point cloud data in terms of rotation. Note that viewpoint variation also leads to a change in self-occlusion. Sensor-level corruptions happen when perceiving with 3D sensors like LiDAR. 
As discussed in prior works~\cite{wu2019squeezesegv2, berger2014state}, sensor-level corruptions can be summarized as 1) dropout noise, where points are missing due to sensor limitations; 
2) spatial inaccuracy, where point positions, object scale, and angle can be wrongly measured; 
3) outliers, which are caused by the structural artifacts in the acquisition process. 
More corruptions could be introduced during postprocessing. 
For example, inaccurate point cloud registration leads to misalignment. 
Background remain and imperfect bounding box are two common corruptions during 3D object scanning.

\begin{figure}[t!]
\vskip 0.in
\begin{center}
\centerline{\includegraphics[width=0.85\linewidth]{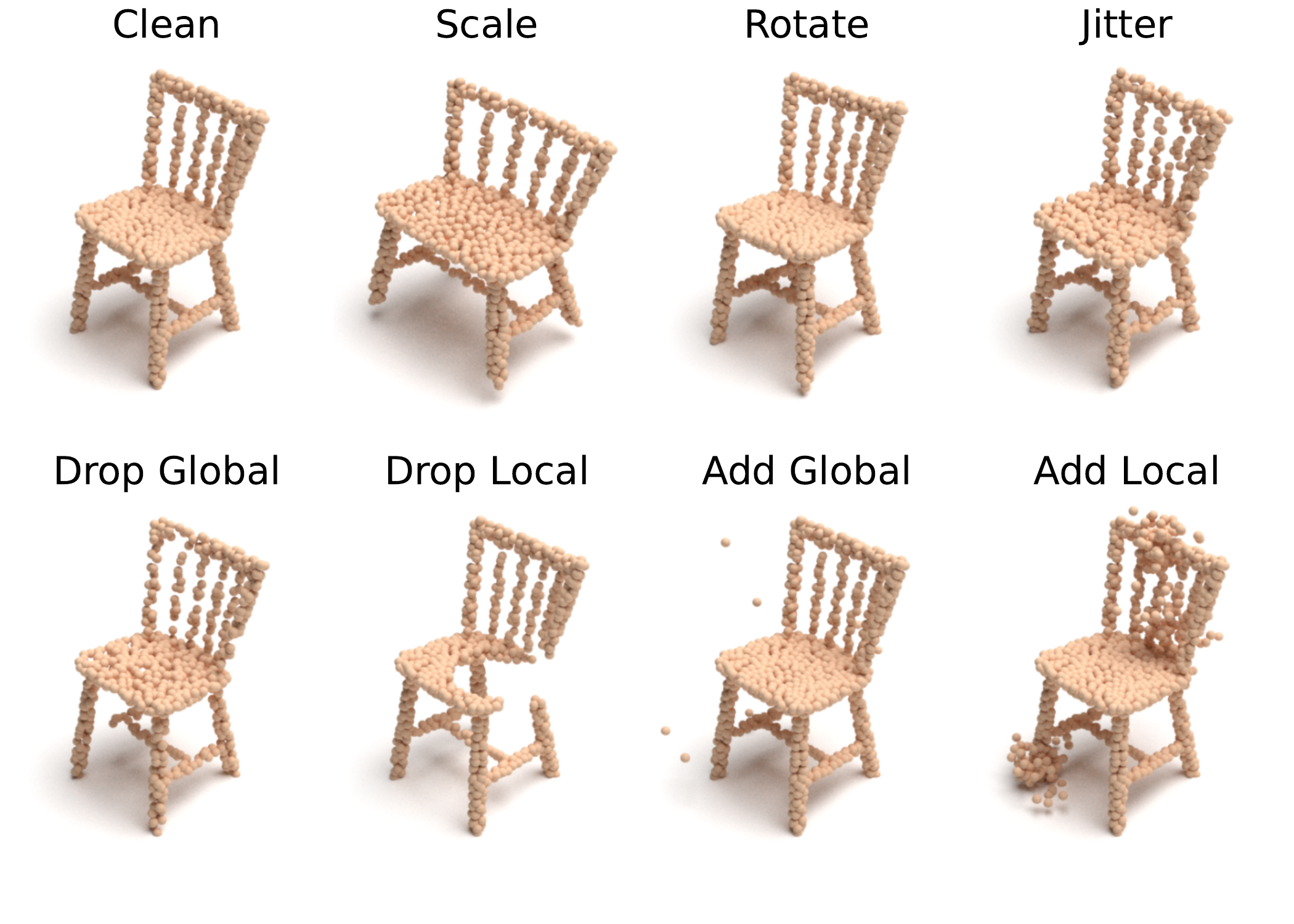}}
\vspace{-7mm}
\caption{Examples of our proposed \emph{\dataset{}}. We corrupt the clean test set of \emph{\dataset{}} using seven types of corruptions with five levels of severity to provide a comprehensive robustness evaluation. Listed examples are from severity level 2. More visualizations on different severity levels can be found in the supplementary material. }
\label{fig:corruptions}
\end{center}
\vskip -0.4in
\end{figure}

However, it is challenging to directly simulate real-world corruptions for the following reasons.
1) Real-world corruptions have a rich variation, \eg, different hardware may have different sensor-level corruptions. 
2) The combination of inter-object occlusion or background remains can be inexhaustive. 
3) Moreover, a few corruptions lead to the same kind of operations to point clouds, \eg, self-occlusion, inter-object occlusion, and cropping error all lead to the missing of a local part of the object. 
To this end, we simplify the corruption taxonomy into seven fundamental atomic corruptions: ``Add Global'', ``Add Local'', ``Drop Global'', ``Drop Local'', ``Rotate'', ``Scale'' and ``Jitter''.
Consequently, each real-world corruption is broken down into a combination of the atomic corruptions, \eg, background remain can be viewed as a combination of ``Add Local'' and ``Add Global''. 

Although the atomic corruptions cannot seamlessly simulate real-world corruptions, they provide a practical solution to achieve controllable empirical study on fundamentally analyzing point cloud classification robustness.
Note that noisy translation and random permutation are not considered in this work, because point cloud normalization and permutation-invariance are two basic properties of recent point cloud classification approaches.

\subsection{\dataset{}: A Robustness Test Suite}
ModelNet40 is one of the most commonly used benchmarks in point cloud classification, and it collects 12,311 CAD models in 40 categories (9,843 for training and 2,468 for testing). 
Most recent point cloud classification methods follow the settings of PointNet~\cite{qi2017pointnet}, which samples 1024 points from each aligned CAD model and then normalizes them into a unit sphere. 
Based on ModelNet40 and the settings by~\cite{qi2017pointnet}, we  
further corrupt the ModelNet40 test set with the aforementioned seven atomic corruptions to establish a comprehensive test-suite \emph{\dataset{}}.
To achieve fair comparisons and meanwhile following the OOD evaluation principle, we use the same training set with ModelNet40.
Similar corruption operations are strictly \textit{not allowed} during training.

The seven atomic corruptions are implemented as follows: ``Scale'' applies a random anisotropic scaling to the point cloud; ``Rotate'' rotates the point cloud by a small angle; ``Jitter'' adds a Gaussian noise to point coordinates; ``Drop Global'' randomly drops points from the point cloud; ``Drop Local'' randomly drops several k-NN clusters from the point cloud; ``Add Global'' adds random points sampled inside a unit sphere; ``Add Local'' expand random points on the point cloud into normally distributed clusters. 
The example corrupted point clouds from \emph{\dataset{}} are shown in \autoref{fig:corruptions}.
In addition, we set different five severity levels for each corruption, based on which we randomly sample from the atomic operations to form a composite corruption test set.
The detailed description and implementation can be found in the appendix. 
Note that we restrict the rotation to small angle variations, as in real-world applications we mostly observe objects from common viewpoints with small variations.
Robustness to arbitrary SO(3) rotations is a specific challenging research topic~\cite{Zhang2019RotationIC, Chen2019ClusterNetDH}, which is out of the scope of this work.

\subsection{Evaluation Metrics}
To normalize the severity of different corruptions, we choose DGCNN, a classic point cloud classification method, as the baseline. 
Inspired by the 2D robustness evaluation metrics~\cite{hendrycks2019bench}, we use mean CE (mCE), as the primary metric.
To compute mCE, we first compute CE:
\begin{align}
    \textrm{CE}_i = \frac{\sum_{l=1}^5 (1 - \textrm{OA}_{i,l})} {\sum_{l=1}^5 (1 - \textrm{OA}^{\textrm{DGCNN}}_{i,l})},
\end{align}
where $\textrm{OA}_{i,l}$ is the overall accuracy on a corrupted test set $i$ at corruption level $l$, $\textrm{OA}^{\textrm{DGCNN}}_{i,l}$ is baseline's overall accuracy
mCE is the average of CE over all seven corruptions:
\begin{align}
    \textrm{mCE} = \frac{1}{N}\sum_{i=1}^N \textrm{CE}_i,
\end{align}
where $N=7$ is the number of corruptions.
We also compute Relative mCE (RmCE), which measures performance drop compared to a clean test set as:
\begin{align}
    &\textrm{RCE}_i = \frac{\sum_{l=1}^5 (\textrm{OA}_{\textrm{Clean}} - \textrm{OA}_{i,l})} {\sum_{l=1}^5 (\textrm{OA}^{\textrm{DGCNN}}_{\textrm{Clean}} -\textrm{OA}^{\textrm{DGCNN}}_{i,l})}, \\
    &\textrm{RmCE} = \frac{1}{N}\sum_{i=1}^N \textrm{RCE}_i,
\end{align}
where $\textrm{OA}_{\textrm{Clean}}$ is the overall accuracy on the clean test set.

\begin{figure}[t!]
\vskip 0.in
\begin{center}
\centerline{\includegraphics[width=\columnwidth]{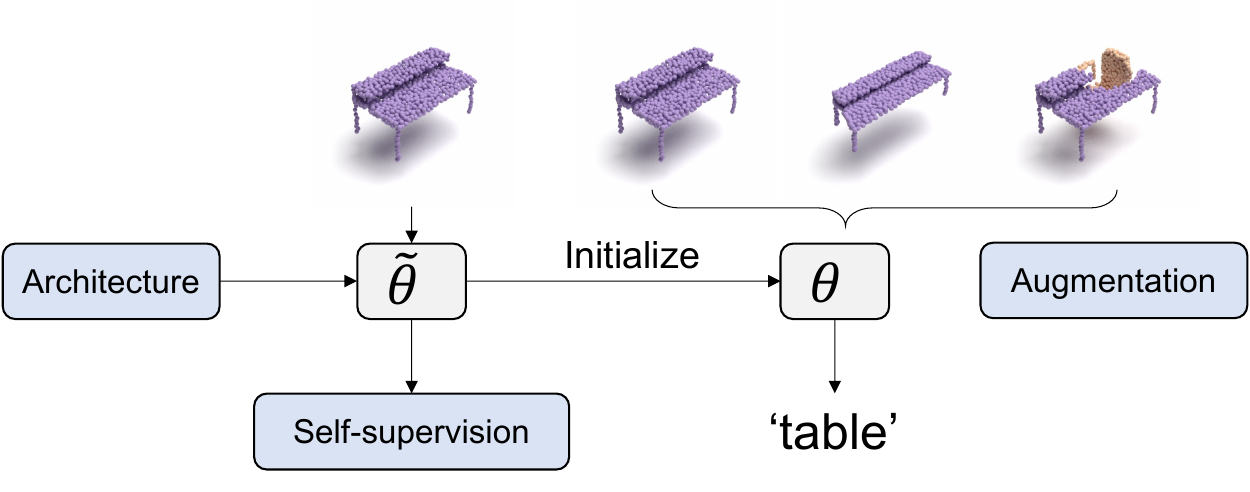}}
\vskip -0.0in
\vspace{-3mm}
\caption{
Robust point cloud classification paradigm. 
Point cloud classification robustness to various corruptions largely depends on three main components, including architecture design, self-supervised pretraining and augmentation methods.
}
\label{fig:overview}
\end{center}
\vskip -0.4in
\end{figure}
\subsection{Evaluation Protocol} 
Because most SoTA methods adopt the \textit{DGCNN protocol}~\cite{goyal2021simpleview}, we also use it as the consistent protocol for the benchmark.
Two conventional augmentations are used during training: 1) random anisotropic scaling in the range [2/3, 3/2]; 2) random translation in the range [-0.2, +0.2]. 
Note that the random scaling ranges for training and testing are not overlapped.
Point cloud sampling is fixed during training, and no voting is used in the inference stage.
For each method, we select the model that performs the best on the clean ModelNet40 test set during evaluation.
We highlight that the same corruptions are not allowed during training to reflect model OOD generalizability.
Following works are recommended to specify augmentations in training when reporting results on \dataset{}.

\section{Systematic Benchmarking}

\noindent\textbf{Implementation Details.}
We benchmark 14 methods in total, covering three key components for robust point cloud classification as shown in \autoref{fig:overview}.
\textbf{Architectures}: PointNet, PointNet++, DGCNN, RSCNN, SimpleView, GDANet, CurveNet, PAConv, 
PCT. 
\textbf{Pretrains}: 
OcCo, 
Point-BERT.
\textbf{Augmentation}: 
PointMixup, 
RSMix, 
PointWOLF.
For PointNet, PointNet++, DGCNN, RSCNN, and SimpleView, we use the pretrained models provided by \citet{goyal2021simpleview}. For CurveNet, GDANet, and PAConv, we use their official pretrained models. 
The rest of the models are trained using their official codes.

\noindent\textbf{Main Results.}
Benchmark results (mCE) are reported in ~\autoref{tab:arch}, ~\autoref{tab:pretrain} and~\autoref{tab:aug} for architechtures, pretrains and augmentations, respectively. 
RmCE and Overall Accuracy are reported in the appendix.
In \autoref{fig:radar}, we sort benchmarked architectures in chronological order and visualize a second-order polynomial fitting results with 50\% confidence interval. We observe that although new architecture's performance are constantly progressing and saturates around 0.94, their mCE performance shows a large variance. 
We also observe that self-supervised pretraining is able to transfer the pretrain signal to the downstream model, but has a mixed effect on the overall performance. Moreover, recent point cloud augmentations can substantially improvement robustness.  

\section{Comprehensive Analysis}
\begin{table}[t]
\caption{Systematic study for architecture design.}
\setlength{\tabcolsep}{3pt}
\centering\scriptsize
\label{tab:systematic}
\begin{tabular}{l|cccc|c}
\toprule
{} & Representation  & \makecell{Local \\ Operations} & \makecell{Advanced \\ Grouping} & Featurizer  & mCE($\downarrow$) \\
\midrule
PointNet & 3D  & No & No & Conventional & 1.422\\
PointNet++ & 3D  & Ball-query & No & Conventional & 1.072\\
DGCNN & 3D  & k-NN  & No & Conventional & 1.000\\
RSCNN & 3D & Ball-query  &  No & Adaptive & 1.130\\
PAConv & 3D & k-NN & No & Adaptive & 1.104\\
CurveNet & 3D & k-NN & Curve & Conventional & 0.927\\
GDANet & 3D & k-NN & Frequency & Conventional & 0.892\\
PCT & 3D & k-NN & No & Self-attention & 0.925\\
SimpleView & 2D & - & - & - & 1.047\\
\midrule
\model{} (Ours)  & 3D & k-NN & Frequency & Self-attention & \textbf{0.863}\\

\bottomrule
\end{tabular}
\end{table}
\begin{figure}[t!]
\vskip 0.in
\begin{center}
\centerline{\includegraphics[width=\columnwidth]{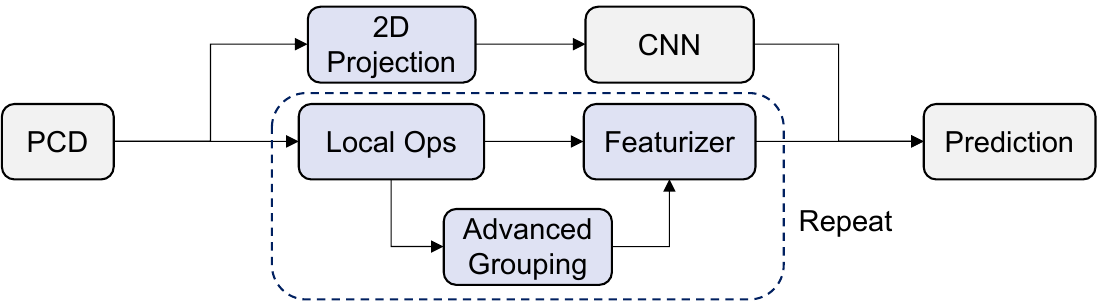}}
\vspace{-3mm}
\caption{Key components in the architecture design. Point cloud data (PCD) repeatedly goes through local operations, advanced grouping, and featurization before being classified. Alternatively, PCD may be projected into multi-view images and processed by traditional CNN-based backbones. This figure means to show how the key components are usually connected, but not to faithfully show every detailed architecture design. 
}
\label{fig:arc_illustration}
\end{center}
\vskip -0.3in
\end{figure}
\begin{table*}
\setlength{\tabcolsep}{11pt}
\centering\scriptsize
\caption{Architectures. Bold: best in column. Underline: second best in column. Blue: best in row. Red: worst in row.}
\label{tab:arch}
\begin{tabular}{lc|c|ccccccc}
\toprule
{} &                               OA $\uparrow$ &                     \textbf{mCE} $\downarrow$&                            Scale &                          Jitter &                          Drop-G &                           Drop-L &                           Add-G &                            Add-L &                           Rotate \\
\midrule
DGCNN~\cite{wang2019dgcnn}      &               \color{black}0.926 &               \color{black}1.000 &                \cellcolor[HTML]{FFFFFF} 1.000 &               \cellcolor[HTML]{FFFFFF} 1.000 &               \cellcolor[HTML]{FFFFFF} 1.000 &                \cellcolor[HTML]{FFFFFF} 1.000 &               \cellcolor[HTML]{FFFFFF} 1.000 &                \cellcolor[HTML]{FFFFFF} 1.000 &                \cellcolor[HTML]{FFFFFF} 1.000 \\
PointNet~\cite{qi2017pointnet}   &               \color{black}0.907 &               \color{black}1.422 &               \color{black}1.266 &    \color{black} \textbf{0.642} &  \cellcolor[HTML]{DFE7FD}  \underline{0.500} &               \color{black}1.072 &                \cellcolor[HTML]{FDE2E4} 2.980 &               \color{black}1.593 &                \cellcolor[HTML]{FFFFFF} 1.902 \\
PointNet++~\cite{qi2017pointnetplusplus} &               \color{black}0.930 &               \color{black}1.072 &               \color{black}0.872 &              \color{black}1.177 &              \color{black}0.641 &                 \cellcolor[HTML]{FDE2E4} 1.802 &     \cellcolor[HTML]{DFE7FD}  \textbf{0.614} &  \color{black} \underline{0.993} &                \cellcolor[HTML]{FFFFFF} 1.405 \\
RSCNN~\cite{liu2019rscnn}      &               \color{black}0.923 &               \color{black}1.130 &               \color{black}1.074 &              \color{black}1.171 &              \color{black}0.806 &                 \cellcolor[HTML]{FDE2E4} 1.517 &  \cellcolor[HTML]{DFE7FD}  \underline{0.712} &               \color{black}1.153 &                \cellcolor[HTML]{FFFFFF} 1.479 \\
SimpleView~\cite{goyal2021simpleview} &     \color{black} \textbf{0.939} &               \color{black}1.047 &               \color{black}0.872 &  \cellcolor[HTML]{DFE7FD}  \underline{0.715} &              \color{black}1.242 &                 \cellcolor[HTML]{FDE2E4} 1.357 &              \color{black}0.983 &     \color{black} \textbf{0.844} &                \cellcolor[HTML]{FFFFFF} 1.316 \\
GDANet~\cite{xu2021gdanet}    &               \color{black}0.934 &  \color{black} \underline{0.892} &     \color{black} \textbf{0.830} &              \color{black}0.839 &               \cellcolor[HTML]{DFE7FD} 0.794 &  \color{black} \underline{0.894} &              \color{black}0.871 &                 \cellcolor[HTML]{FDE2E4} 1.036 &                \cellcolor[HTML]{FFFFFF} 0.981 \\
CurveNet~\cite{xiang2021curvenet}   &  \color{black} \underline{0.938} &               \color{black}0.927 &               \color{black}0.872 &              \color{black}0.725 &               \cellcolor[HTML]{DFE7FD} 0.710 &                \cellcolor[HTML]{FFFFFF} 1.024 &                \cellcolor[HTML]{FDE2E4} 1.346 &               \color{black}1.000 &     \color{black} \textbf{0.809} \\
PAConv~\cite{xu2021paconv}     &               \color{black}0.936 &               \color{black}1.104 &                \cellcolor[HTML]{DFE7FD} 0.904 &                \cellcolor[HTML]{FDE2E4} 1.465 &              \color{black}1.000 &               \color{black}1.005 &              \color{black}1.085 &                \cellcolor[HTML]{FFFFFF} 1.298 &  \color{black} \underline{0.967} \\
PCT~\cite{guo2020pct}        &               \color{black}0.930 &               \color{black}0.925 &               \color{black}0.872 &              \color{black}0.870 &               \cellcolor[HTML]{DFE7FD} 0.528 &               \color{black}1.000 &              \color{black}0.780 &                 \cellcolor[HTML]{FDE2E4} 1.385 &                \cellcolor[HTML]{FFFFFF} 1.042 \\
\midrule \model{} (Ours)  &               \color{black}0.930 &     \color{black} \textbf{0.863} &  \color{black} \underline{0.840} &              \color{black}0.892 &     \cellcolor[HTML]{DFE7FD}  \textbf{0.492} &     \color{black} \textbf{0.797} &              \color{black}0.929 &                \cellcolor[HTML]{FFFFFF} 1.011 &                 \cellcolor[HTML]{FDE2E4} 1.079 \\
\bottomrule
\end{tabular}
\vskip -0.1in
\end{table*}

\begin{table*}
\setlength{\tabcolsep}{11pt}
\centering\scriptsize
\caption{Pretrain. $\dagger$: randomly initialized. Bold: best in column. Underline: second best in column. Blue: best in row. Red: worst in row.}
\label{tab:pretrain}
\begin{tabular}{lc|c|ccccccc}
\toprule
{} &                               OA $\uparrow$ &                     \textbf{mCE}  $\downarrow$&                         Scale &                           Jitter &                          Drop-G &                           Drop-L &                           Add-G &                           Add-L &                          Rotate \\
\midrule
DGCNN~\cite{wang2019dgcnn}       &     \color{black} \textbf{0.926} &     \color{black} \textbf{1.000} &             \cellcolor[HTML]{FFFFFF}1.000 &                \cellcolor[HTML]{FFFFFF}1.000 &               \cellcolor[HTML]{FFFFFF}1.000 &      \cellcolor[HTML]{FFFFFF} \textbf{1.000} &     \cellcolor[HTML]{FFFFFF} \textbf{1.000} &  \cellcolor[HTML]{FFFFFF} \underline{1.000} &  \cellcolor[HTML]{FFFFFF} \underline{1.000} \\
+OcCo~\cite{wang2021occo}  &  \color{black} \underline{0.922} &  \color{black} \underline{1.047} &              \cellcolor[HTML]{FDE2E4}1.606 &      \cellcolor[HTML]{DFE7FD} \textbf{0.652} &              \color{black}0.903 &  \color{black} \underline{1.039} &  \cellcolor[HTML]{FFFFFF} \underline{1.444} &    \color{black} \textbf{0.847} &    \color{black} \textbf{0.837} \\
\midrule
Point-BERT\textsuperscript{$\dagger$} &               \color{black}0.919 &               \color{black}1.317 &  \color{black} \textbf{0.936} &  \color{black} \underline{0.987} &  \cellcolor[HTML]{DFE7FD} \underline{0.899} &               \color{black}1.295 &                \cellcolor[HTML]{FDE2E4}2.336 &              \color{black}1.360 &               \cellcolor[HTML]{FFFFFF}1.409 \\
+Point-BERT~\cite{yu2021pointbert}  &  \color{black} \underline{0.922} &               \color{black}1.248 &  \color{black} \textbf{0.936} &               \color{black}1.259 &     \cellcolor[HTML]{DFE7FD} \textbf{0.690} &               \color{black}1.150 &                \cellcolor[HTML]{FDE2E4}1.932 &               \cellcolor[HTML]{FFFFFF}1.440 &              \color{black}1.326 \\
\bottomrule
\end{tabular}
\end{table*}

\subsection{Architecture Design} 
We analyze four key components of point cloud classifier architectures: local operations, advanced grouping, featurizer, and representation dimension, as illustrated in \autoref{fig:arc_illustration}. The design choices of recent classifier architectures are summarized in \autoref{tab:systematic}. When analyzing a specific component, we group all methods that utilize the component. Since design choices are not rigorously controlled variables in the analysis, we visualize the 95\% confidence interval together with the mean value in the bar charts, and only low variance results are considered in our conclusion. 
Furthermore, to empirically verify our conclusion, we build a new architecture, \model{}, strictly following the conclusions.

\begin{figure*}
\begin{subfigure}{0.67\columnwidth}
  \centering
  \includegraphics[width=\linewidth]{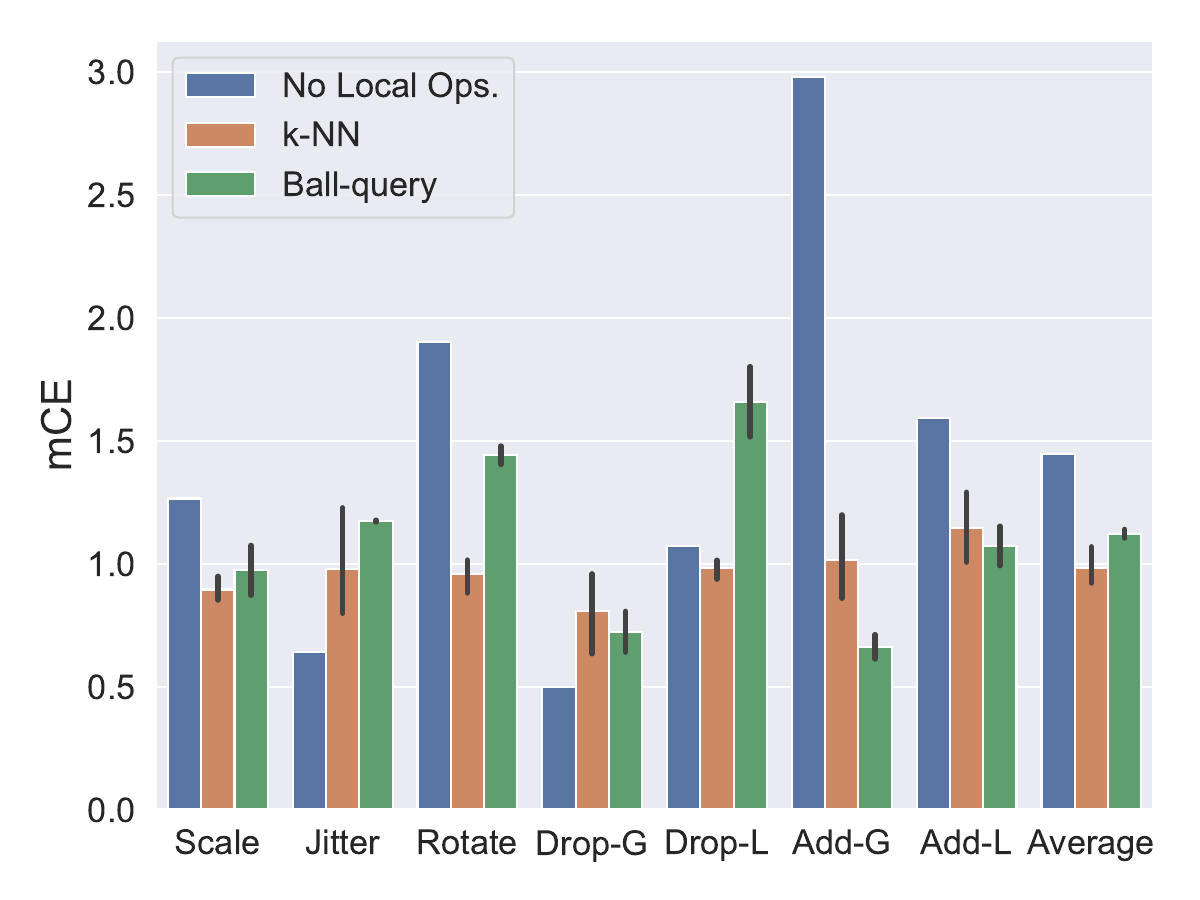}
   \vskip -0.1in
  \caption{Local Operations}
  \label{fig:Local operations}
\end{subfigure}%
\begin{subfigure}{0.67\columnwidth}
  \centering
  \includegraphics[width=\linewidth]{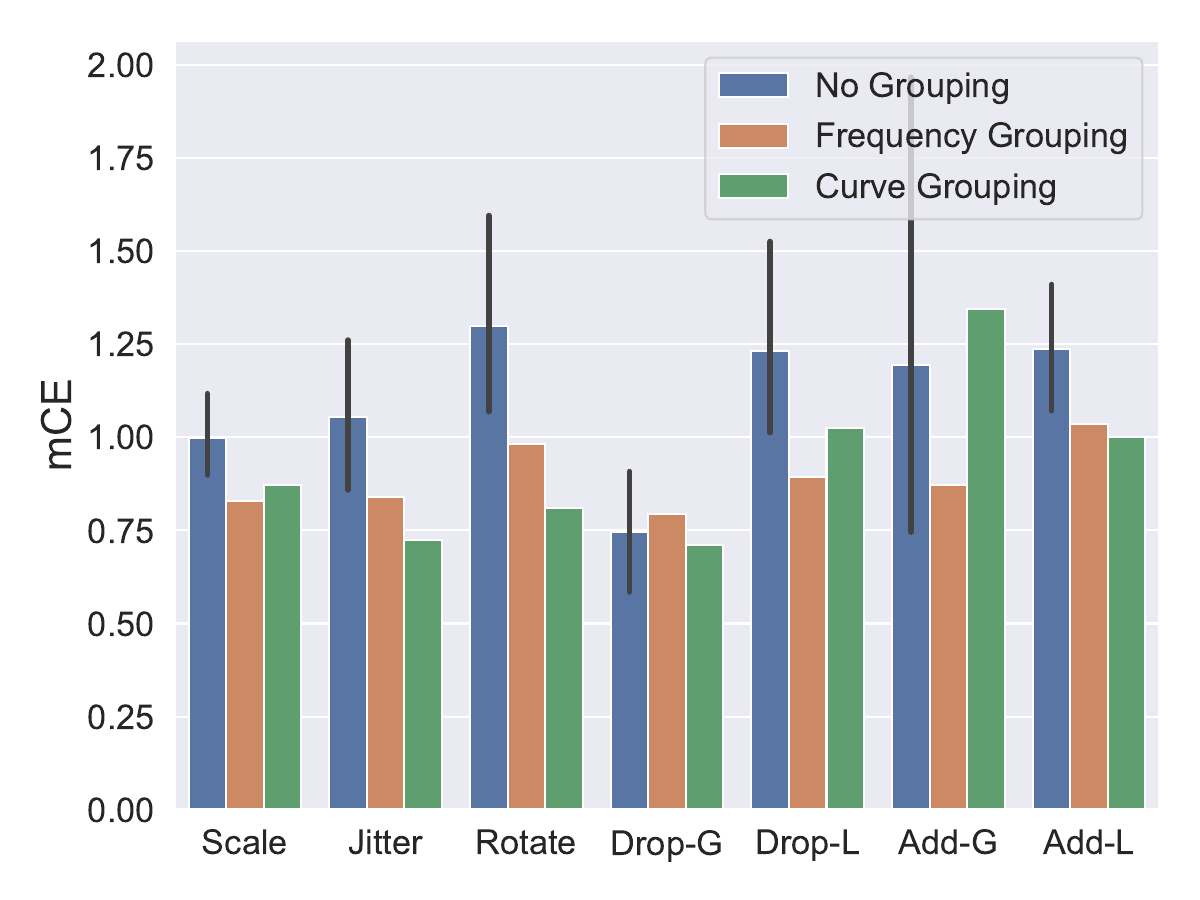}
    \vskip -0.1in
  \caption{Advanced Grouping}
  \label{fig:Advanced Grouping}
\end{subfigure}
\begin{subfigure}{0.67\columnwidth}
  \centering
  \includegraphics[width=\linewidth]{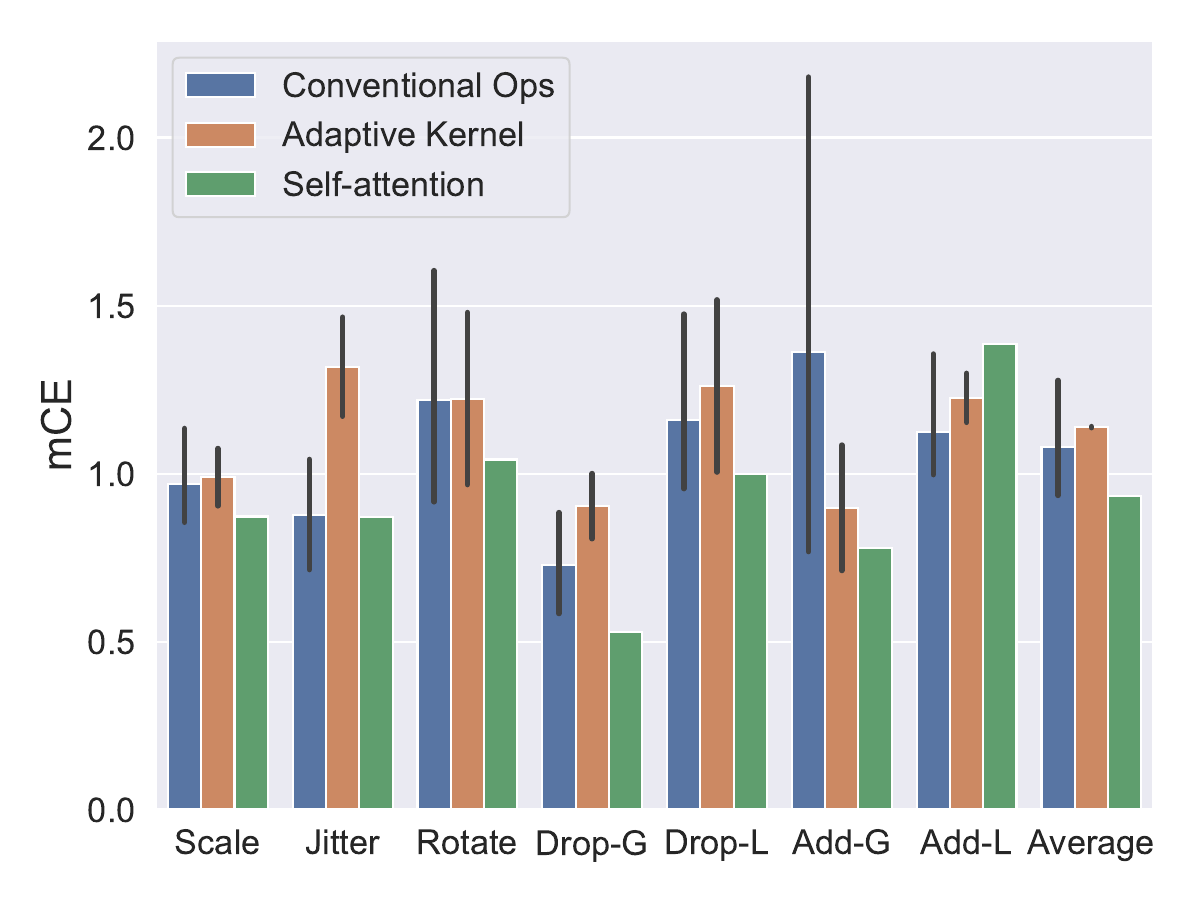}
     \vskip -0.1in
  \caption{Featurizer}
  \label{fig:Featurizer}
\end{subfigure}

\begin{subfigure}{0.67\columnwidth}
  \centering
  \includegraphics[width=\linewidth]{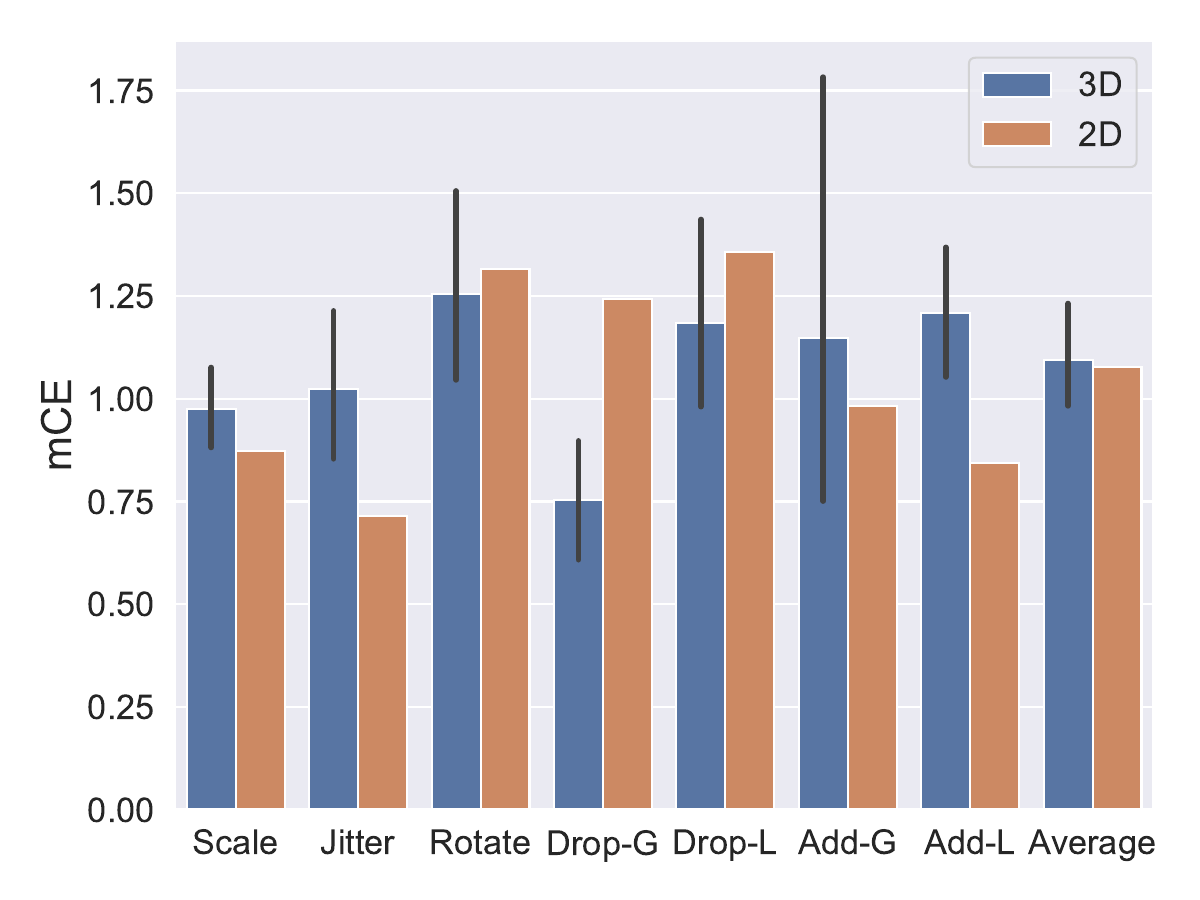}
     \vskip -0.1in
  \caption{2D v.s. 3D}
  \label{fig:2D v.s. 3D}
\end{subfigure}%
\begin{subfigure}{0.67\columnwidth}
  \centering
  \includegraphics[width=\linewidth]{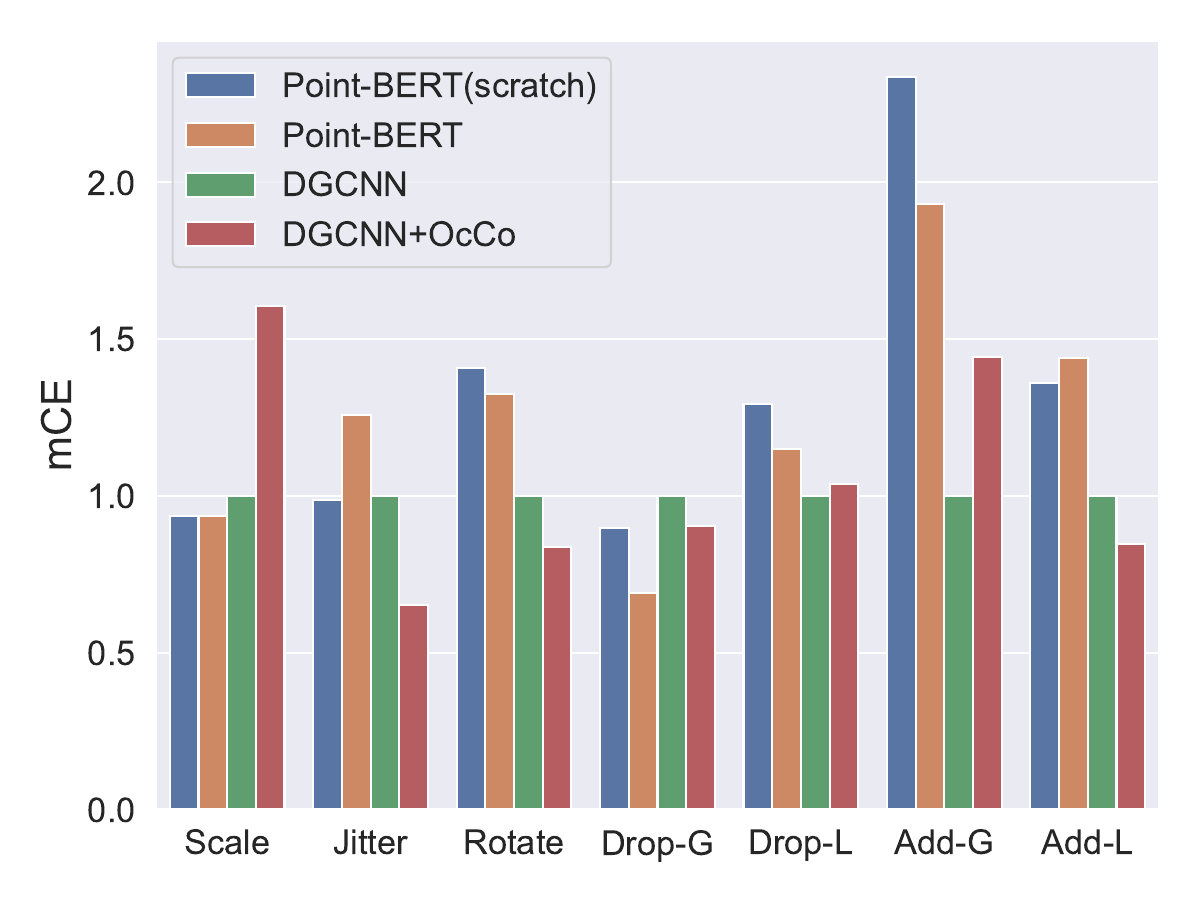}
     \vskip -0.1in
  \caption{Pretrain}
  \label{fig:Pretrain}
\end{subfigure}
\begin{subfigure}{0.67\columnwidth}
  \centering
  \includegraphics[width=\linewidth]{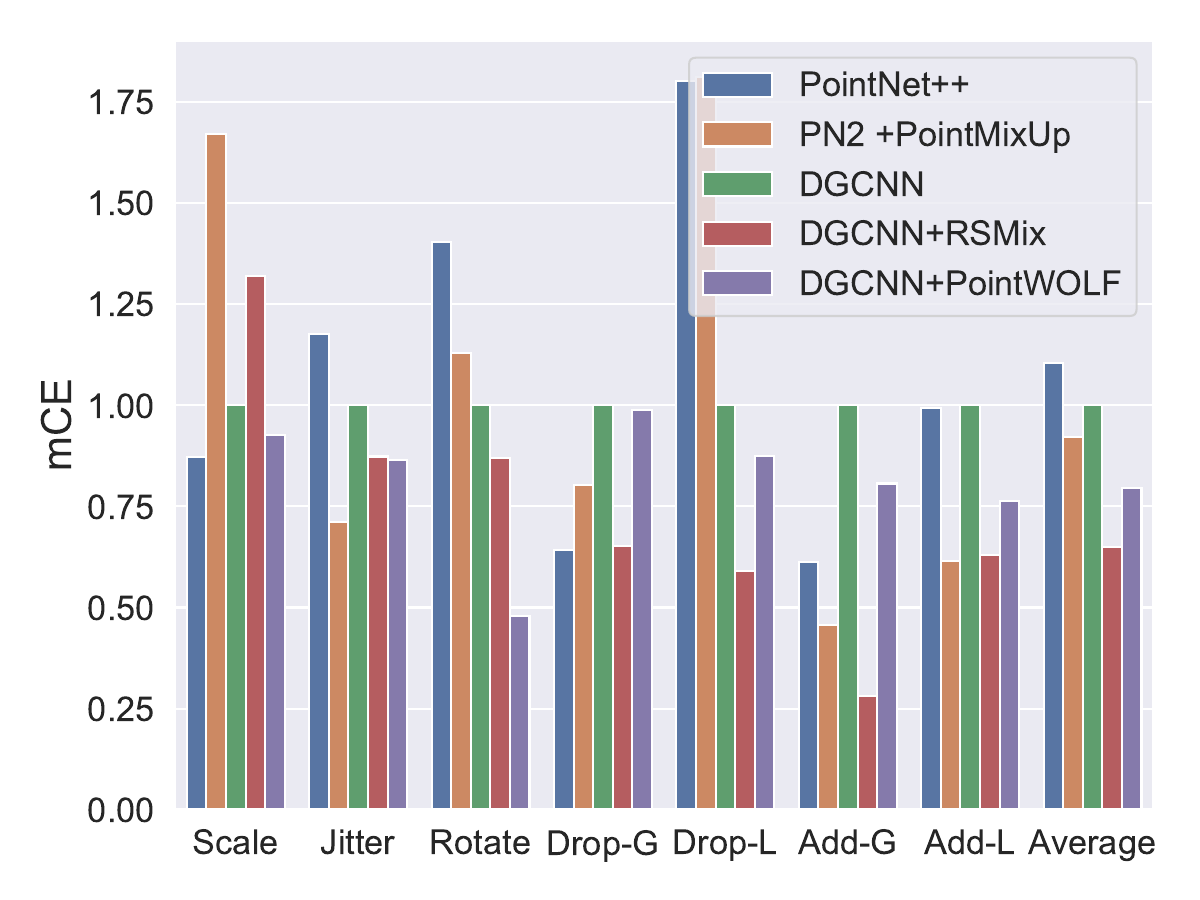}
     \vskip -0.1in
  \caption{Augmentation}
  \label{fig:Augmentation}
\end{subfigure}
\vspace{-3mm}
\caption{Analysis on different architecture designs, pretrain strategies and augmentation strategies' effect to classifier's performance under different corruptions. "-G": Global. "-L": Local.}
\end{figure*}

\noindent\textbf{Local Operations.}
We compare the robustness of different local aggregations, including no local operations, k-NN, and ball-query. As shown in \autoref{fig:Local operations}, the exploitation of the point cloud locality is a key component to robustness.
Without local aggregations, PointNet (shown as ``No Local Ops.'') has the highest mCE.
Considering each corruption individually, PointNet is on the two extremes: it shows the best robustness to ``Jitter'' and ``Drop-G'', meanwhile being one of the worst methods for the rest corruptions. 
Local operations target to encode informative representations by exploiting local geometric features. 
Ball-query randomly samples neighboring points in a predefined radius, while k-NN focuses on nearest neighboring points.
Generally, \emph{k-NN performs better than ball-query in the benchmark,} especially for ``Drop-L''.
The reason is that points surrounding the dropped local part will lose its neighbors in ball-query due to its fixed searching radius, but k-NN will choose neighbors from the remaining points. 
However, ball-query shows the advantage over k-NN in ``Add-G'', since, for a point on the object, outliers are less likely to fall in the query ball than to be its nearest neighbors.

\noindent\textbf{Advanced Grouping.}
Recent methods design advanced grouping techniques, such as Frequency Grouping~\cite{xu2021gdanet} and Curve Grouping~\cite{xiang2021curvenet}, to introduce structural prior into architecture design. Frequency grouping uses a graph high-pass filter~\cite{sandryhaila2014discrete, ortega2018graph} to group point features in the frequency domain. 
Curve grouping forms a curve-like point set $\{P_1, P_2, ... P_N\}$ by walking from $P_i$ to $P_{i+1}$ following a learnable policy $\pi$. As shown in \autoref{fig:Advanced Grouping}, we observe that \emph{both grouping techniques improve model robustness by a clear margin.} 
The idea of frequency grouping aligns with the observations in~\cite{yin2019fourier}: there is a trade-off between model robustness to low-frequency corruptions and high-frequency corruptions. By viewing local-grouped features as low-frequency features and curve-grouped feature as high-frequency features, the robustness gain can be again interpreted from a frequency perspective. Nonetheless, it is noteworthy that advanced grouping is more time-consuming during both training and testing.

\noindent\textbf{Featurizer.} 
We refer conventional operators to shared MLPs and convolutional layers, which are common building blocks for point cloud models. 
Recent works explore various advanced feature processing methods, such as adaptive kernels and self-attention operations. RSCNN~\cite{liu2019rscnn} and PAConv~\cite{xu2021paconv} design adaptive kernels whose weights change with low-level features like spatial coordinates and surface normals. 
Based on self-attention, PCT~\cite{guo2020pct} proposes the offset-attention operation, which achieves impressive performance for point cloud analysis.
Despite the success of RSCNN and PAConv on clean point cloud classifications, they tend to be more sensitive to corruptions than conventional operators in our experiments shown in \autoref{fig:Featurizer}. Data corruption exacerbates through data-dependent kernels. 
Compared to conventional operators, \emph{self-attention operations improve classifier robustness in several aspects,} particularly in ``Drop-G''. 
We speculate that its robustness gains to ``Drop-G'' come from its ability to understand non-local relations from the global perspective. Note that Point-BERT~\cite{yu2021pointbert} also introduces an self-attention-based architecture. However, it includes a fixed tokenizer that is trained on pretext tasks, which could be the bottleneck for its robustness performance.
Therefore, we do not include the randomly initialized Point-BERT result in the architecture analysis. 

\noindent\textbf{2D vs. 3D Representation.}
A few methods~\cite{qi2016volumetric, goyal2021simpleview} first project 3D shapes to 2D frames from different viewpoints, and then use 2D classifiers for recognizing 3D points.
The recently proposed projection-based method,  SimpleView~\cite{goyal2021simpleview} performs surprisingly well on clean 3D point clouds.
In our experiments shown in~\autoref{fig:2D v.s. 3D}, \emph{projecting 3D points to 2D images brought mixed effects to classification.} The projection significantly reduces the effect of ``Jitter'' and ``Add-L'', but suffers a lot from point scarcity, particularly ``Drop-G''. 
This is consistent with human visual perception, as 
it is challenging for human vision to recognize the shape from point projections, especially for sparse and noisy points without texture information.
Adding more observations from different perspectives might improve 2D perception accuracy, while extra efforts are required.
In a nutshell, we think 3D cues are more straight-forward and preferable for building a robust point cloud classifier.

\subsection{Self-supervised Pretraining}
Recently, various self-supervised pretrain methods have been proposed for point cloud classification models, such as Point-BERT~\cite{yu2021pointbert} and OcCo~\cite{wang2021occo}.
We study their robustness against corruptions in~\autoref{fig:Pretrain}, which reveals that \emph{pretrain signals can be transferred}, and hence benefiting classification under specific corruptions. 
During self-supervised pretrain, Point-BERT first drops points using the block-wise masking strategy and then reconstructs the missing points based on the rest points. Interestingly, models finetuned on Point-BERT pretrain show better classification robustness when local part is missing. OcCo employs a similar reconstruction pretrain task, but with a different masking strategy. By observing from different camera viewpoints, OcCo masks the points that are self-occluded. Meanwhile, point clouds are also rotated with different camera angles. Consequently, the OcCo pretrained models are significantly more robust to rotation perturbations. Moreover, OcCo also improves the robustness to ``Jitter'' and ``Add-L''.

\begin{table*}
\setlength{\tabcolsep}{10pt}
\centering\scriptsize
\caption{Augmentation. Bold: best in column. Underline: second best in column. Blue: best in row. Red: worst in row.}
\label{tab:aug}
\begin{tabular}{lc|c|ccccccc}
\toprule
{} &                               OA  $\uparrow$&                     \textbf{mCE} $\downarrow$ &                           Scale &                          Jitter &                           Drop-G &                           Drop-L &                           Add-G &                            Add-L &                          Rotate \\
\midrule
DGCNN~\cite{wang2019dgcnn}           &               \color{black}0.926 &               \color{black}1.000 &               \cellcolor[HTML]{FFFFFF} 1.000 &               \cellcolor[HTML]{FFFFFF} 1.000 &                \cellcolor[HTML]{FFFFFF} 1.000 &                \cellcolor[HTML]{FFFFFF} 1.000 &               \cellcolor[HTML]{FFFFFF} 1.000 &                \cellcolor[HTML]{FFFFFF} 1.000 &               \cellcolor[HTML]{FFFFFF} 1.000 \\
+PointWOLF~\cite{kim2021pointwolf} &               \color{black}0.926 &               \color{black}0.814 &  \cellcolor[HTML]{FFFFFF}  \underline{0.926} &              \color{black}0.864 &                 \cellcolor[HTML]{FDE2E4} 0.988 &               \color{black}0.874 &              \color{black}0.807 &               \color{black}0.764 &  \cellcolor[HTML]{DFE7FD}  \underline{0.479} \\
+RSMix~\cite{lee2021rsmix}     &  \color{black} \underline{0.930} &  \color{black} \underline{0.745} &                \cellcolor[HTML]{FDE2E4} 1.319 &               \cellcolor[HTML]{FFFFFF} 0.873 &  \color{black} \underline{0.653} &  \color{black} \underline{0.589} &     \cellcolor[HTML]{DFE7FD}  \textbf{0.281} &               \color{black}0.629 &              \color{black}0.870 \\
+\aug{} (Ours)   &     \color{black} \textbf{0.932} &     \color{black} \textbf{0.590} &                \cellcolor[HTML]{FDE2E4} 0.989 &  \cellcolor[HTML]{FFFFFF}  \underline{0.715} &               \color{black}0.698 &     \color{black} \textbf{0.575} &  \cellcolor[HTML]{DFE7FD}  \underline{0.285} &     \color{black} \textbf{0.415} &    \color{black} \textbf{0.451} \\
\midrule
PointNet++~\cite{qi2017pointnetplusplus}      &  \color{black} \underline{0.930} &               \color{black}1.072 &    \color{black} \textbf{0.872} &              \color{black}1.177 &     \color{black} \textbf{0.641} &                 \cellcolor[HTML]{FDE2E4} 1.802 &               \cellcolor[HTML]{DFE7FD} 0.614 &               \color{black}0.993 &               \cellcolor[HTML]{FFFFFF} 1.405 \\
+PointMixUp~\cite{chen2020pointmixup} &               \color{black}0.915 &               \color{black}1.028 &               \cellcolor[HTML]{FFFFFF} 1.670 &    \color{black} \textbf{0.712} &               \color{black}0.802 &                 \cellcolor[HTML]{FDE2E4} 1.812 &               \cellcolor[HTML]{DFE7FD} 0.458 &  \color{black} \underline{0.615} &              \color{black}1.130 \\
\bottomrule
\end{tabular}
\end{table*}

\begin{table*}
\setlength{\tabcolsep}{11pt}
\centering\scriptsize
\caption{Results of combining \aug{} with different architectures. Bold: best in column. Underline: second best in column. Blue: best in row. Red: worst in row.}
\label{tab:wolfmix}
\begin{tabular}{lc|c|ccccccc}
\toprule
{} &                            OA $\uparrow$ &                     \textbf{mCE}  $\downarrow$&                            Scale &                           Jitter &                          Drop-G &                           Drop-L &                           Add-G &                         Add-L &                           Rotate \\
\midrule
DGCNN~\cite{wang2019dgcnn}                           &            \color{black}0.926 &               \color{black}1.000 &                1.000 &                1.000 &              1.000 &                1.000 &               1.000 &             1.000 &                1.000 \\
+\aug{}                   &            \color{black}0.932 &  \color{black} 0.590 &                 \cellcolor[HTML]{FDE2E4}0.989 &                \cellcolor[HTML]{FFFFFF}0.715 &              \color{black}0.698 &  \color{black} 0.575 &     \cellcolor[HTML]{DFE7FD} \textbf{0.285} &  \color{black} \textbf{0.415} &  \color{black} \underline{0.451} \\
\midrule
PointNet~\cite{qi2017pointnet}                        &            \color{black}0.907 &               \color{black}1.422 &               \color{black}1.266 &  \color{black} \underline{0.642} &               \cellcolor[HTML]{DFE7FD}0.500 &               \color{black}1.072 &                \cellcolor[HTML]{FDE2E4}2.980 &            \color{black}1.593 &                \cellcolor[HTML]{FFFFFF}1.902 \\
+\aug{}                &            \color{black}0.884 &               \color{black}1.180 &                \cellcolor[HTML]{FFFFFF}2.117 &      \cellcolor[HTML]{DFE7FD} \textbf{0.475} &              \color{black}0.577 &               \color{black}1.082 &                \cellcolor[HTML]{FDE2E4}2.227 &            \color{black}0.702 &               \color{black}1.079 \\
\midrule
PCT~\cite{guo2020pct}                             &            \color{black}0.930 &               \color{black}0.925 &               \color{black}0.872 &               \color{black}0.870 &               \cellcolor[HTML]{DFE7FD}0.528 &               \color{black}1.000 &              \color{black}0.780 &              \cellcolor[HTML]{FDE2E4}1.385 &                \cellcolor[HTML]{FFFFFF}1.042 \\
+\aug{}                     &            \color{black}\textbf{0.934} & \color{black}\underline{0.574} &               \cellcolor[HTML]{FDE2E4}1.000 &               \color{black}0.854 &               \color{black}\textbf{0.379} &    \color{black} \textbf{0.493} &                \cellcolor[HTML]{DFE7FD}\underline{0.298} &              \color{black}0.505 &                 \color{black}0.488 \\
\midrule
GDANet~\cite{xu2021gdanet}                          &  \color{black} \textbf{0.934} &               \color{black}0.892 &     \color{black} \textbf{0.830} &               \color{black}0.839 &               \cellcolor[HTML]{DFE7FD}0.794 &               \color{black}0.894 &              \color{black}0.871 &              \cellcolor[HTML]{FDE2E4}1.036 &                \cellcolor[HTML]{FFFFFF}0.981 \\
+\aug{}                     &  \color{black} \textbf{0.934} &     \color{black} \textbf{0.571} &                 \cellcolor[HTML]{FDE2E4}0.904 &                \cellcolor[HTML]{FFFFFF}0.883 &              \color{black}0.532 &     \color{black} 0.551 &  \cellcolor[HTML]{DFE7FD} {0.305} &  \color{black} \textbf{0.415} &     \color{black} \textbf{0.409} \\
\midrule
\model{}                       &            \color{black}0.930 &               \color{black}0.863 &  \color{black} \underline{0.840} &               \color{black}0.892 &  \cellcolor[HTML]{DFE7FD} 0.492 &               \color{black}0.797 &              \color{black}0.929 &             \cellcolor[HTML]{FFFFFF}1.011 &                 \cellcolor[HTML]{FDE2E4}1.079 \\
+\aug{}               &            \color{black}\underline{0.933} &               \color{black}0.601 &               \cellcolor[HTML]{FDE2E4}1.011 &                \cellcolor[HTML]{FFFFFF}0.968 &              \color{black}\underline{0.423} &               \color{black}\underline{0.512} &                \cellcolor[HTML]{DFE7FD}0.332 &            \color{black}\underline{0.480} &                0.479 \\
\bottomrule
\end{tabular}
\end{table*}

\subsection{Augmentation Method}
Following the principle of OOD evaluation, the corruptions should not be used as augmentations during training, and therefore we choose mixing and deformation augmentations.
As shown in~\autoref{fig:Augmentation},
\emph{mixing and deformation augmentations can bring significant improvements to model robustness.} 
PointMixUp~\cite{chen2020pointmixup} and RSMix~\cite{lee2021rsmix} are two mix strategies. 
Similar to MixUp~\cite{zhang2018mixup} in 2D augmentation, PointMixup mixes two point clouds using shortest-path interpolation. 
Similar to CutMix~\cite{yun2019cutmix} in 2D augmentation, RSMix mixes two point clouds using rigid transformation. 
Both mix strategies substantially reduce CE on corruptions including ``Add-G'', ``Add-L'', ``Rotate'' and ``Jitter''. 
However, an unexpected side effect of the mix strategies is that classifiers become more vulnerable to scaling effects. 
By non-rigidly deforming local parts of an object, 
PointWOLF~\cite{kim2021pointwolf} enrich the data variation, which constantly improves classifier robustness on all evaluated corruptions.

\section{Boosting Corruption Robustness}
Based on the above observations, we propose to improve point cloud classifier robustness in the following ways.

\noindent\textbf{\model{}: A Robust Point Cloud Classifier.}
Following the conclusions in the architecture analysis, we construct \model{} using 3D representation, k-NN, frequency grouping and self-attention. The detailed architecture is shown in the appendix. As reported in~\autoref{tab:arch}, \model{} achieves the best mCE compared to all SoTA methods. 
The success of \model{} empirically verifies our conclusions on the architecture design choices, and it could serve as a strong baseline for future robustness research. 
The implementation details are provided in the appendix.

\begin{figure}[t!]
\vskip 0.in
\begin{center}
\centerline{\includegraphics[width=1\columnwidth]{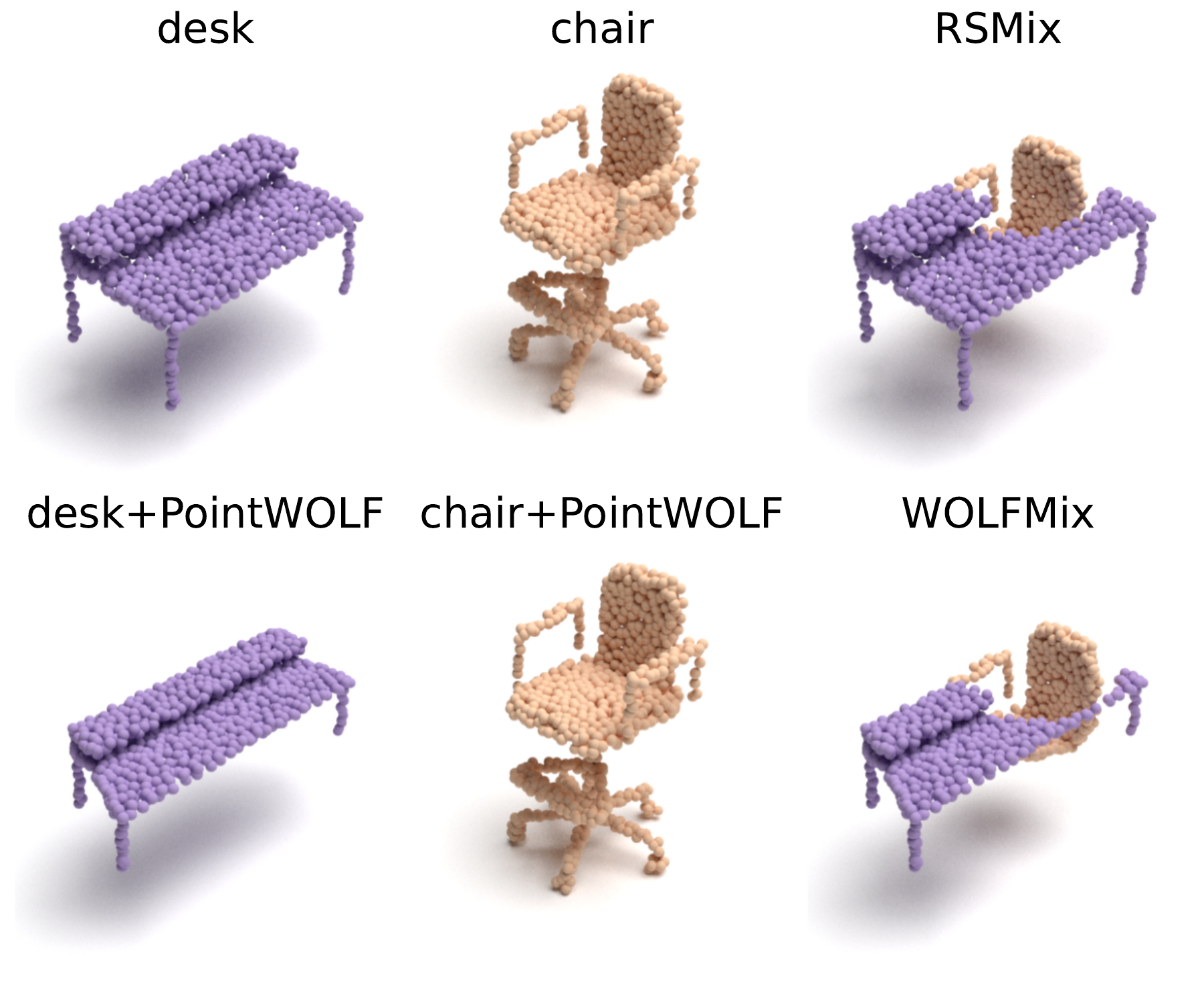}}
\vskip -0.0in
\vspace{-8mm}
\caption{Illustration of the proposed WOLFMix augmentation. Point clouds are first locally deformed and then rigidly mixed. Ground truth labels are mixed accordingly. }
\label{fig:wolfmix}
\end{center}
\vskip -0.4in
\end{figure}

\noindent\textbf{\aug{}: A Strong Augmentation Strategy.}
We design \aug{} upon two powerful augmentation strategies, PointWOLF and RSMix.
During training, \aug{} first deforms the object, and then rigidly mixes the two deformed objects together. 
Ground-truth labels are mixed accordingly. 
We show an illustration of \aug{} in~\autoref{fig:wolfmix}. 
By taking advantage of both rigid and non-rigid transformations, \aug{} brings substantial robustness gain over standalone PointWOLF and RSMix in~\autoref{tab:aug}. 
Implementation details can be found in the appendix.

\noindent\textbf{Synergy between Architecture and Augmentation.} 
We observe that augmentation techniques do not equally transfer to different architectures. 
\autoref{tab:wolfmix} shows that the improvement by \aug{} on corruption robustness varies with different models.
Although \model{} achieves the lowest standalone mCE, 
its improvements by \aug{} are less than \aug{} for DGCNN, PCT and GDANet.
PointNet enjoys limited robustness gain as well. 
Hence, we speculate that there is a capacity upper bound to corruptions for each architecture. 
Future classification robustness research is suggested to study: 1) standalone robustness for architecture and augmentations independently; and 2) their synergy in between.
Furthermore, we identify that training GDANet with \aug{} achieves the best robustness in all existing methods, with an impressive 0.571 mCE.

\section{Conclusion}

In this work, we establish a comprehensive test suite \emph{\dataset{}} for robust point cloud classification under corruptions. 
We systematically benchmarked and analyzed representative point cloud classification methods.
By analyzing benchmark results, we propose two effective strategies, \model{} and \aug{}, for improving robustness. 
As the SoTA methods for point cloud classification on clean data are becoming less robust to random real-world corruptions, we highly encourage future research to focus on classification robustness so as to benefit real applications.

\section{Acknowledgment}
This work is supported by the National Research Foundation, Singapore under its AI Singapore Programme (AISG Award No: AISG2-PhD-2021-08-018), NTU NAP, MOE AcRF Tier 2 (T2EP20221-0033), and under the RIE2020 Industry Alignment Fund – Industry Collaboration Projects (IAF-ICP) Funding Initiative, as well as cash and in-kind contribution from the industry partner(s).

\clearpage
\bibliography{ref}

\begin{thebibliography}{47}
\providecommand{\natexlab}[1]{#1}
\providecommand{\url}[1]{\texttt{#1}}
\expandafter\ifx\csname urlstyle\endcsname\relax
  \providecommand{\doi}[1]{doi: #1}\else
  \providecommand{\doi}{doi: \begingroup \urlstyle{rm}\Url}\fi

\bibitem[Ahmadyan et~al.(2021)Ahmadyan, Zhang, Ablavatski, Wei, and
  Grundmann]{ahmadyan2021objectron}
Ahmadyan, A., Zhang, L., Ablavatski, A., Wei, J., and Grundmann, M.
\newblock Objectron: A large scale dataset of object-centric videos in the wild
  with pose annotations.
\newblock In \emph{Proceedings of the IEEE/CVF Conference on Computer Vision
  and Pattern Recognition}, pp.\  7822--7831, 2021.

\bibitem[Barbu et~al.(2019)Barbu, Mayo, Alverio, Luo, Wang, Gutfreund,
  Tenenbaum, and Katz]{barbu2019objectnet}
Barbu, A., Mayo, D., Alverio, J., Luo, W., Wang, C., Gutfreund, D., Tenenbaum,
  J., and Katz, B.
\newblock Objectnet: {A} large-scale bias-controlled dataset for pushing the
  limits of object recognition models.
\newblock In \emph{NeurIPS}, pp.\  9448--9458, 2019.

\bibitem[Berger et~al.(2014)Berger, Tagliasacchi, Seversky, Alliez, Levine,
  Sharf, and Silva]{berger2014state}
Berger, M., Tagliasacchi, A., Seversky, L., Alliez, P., Levine, J., Sharf, A.,
  and Silva, C.
\newblock State of the art in surface reconstruction from point clouds.
\newblock In \emph{Eurographics 2014-State of the Art Reports}, volume~1, pp.\
  161--185, 2014.

\bibitem[Chen et~al.(2019)Chen, Li, Xu, Chen, Wang, and
  Lin]{Chen2019ClusterNetDH}
Chen, C., Li, G., Xu, R., Chen, T., Wang, M., and Lin, L.
\newblock Clusternet: Deep hierarchical cluster network with rigorously
  rotation-invariant representation for point cloud analysis.
\newblock \emph{2019 IEEE/CVF Conference on Computer Vision and Pattern
  Recognition (CVPR)}, pp.\  4989--4997, 2019.

\bibitem[Chen et~al.(2020)Chen, Hu, Gavves, Mensink, Mettes, Yang, and
  Snoek]{chen2020pointmixup}
Chen, Y., Hu, V.~T., Gavves, E., Mensink, T., Mettes, P., Yang, P., and Snoek,
  C. G.~M.
\newblock Pointmixup: Augmentation for point clouds.
\newblock In \emph{{ECCV} {(3)}}, volume 12348 of \emph{Lecture Notes in
  Computer Science}, pp.\  330--345. Springer, 2020.

\bibitem[Deng et~al.(2021)Deng, Litany, Duan, Poulenard, Tagliasacchi, and
  Guibas]{deng2021vector}
Deng, C., Litany, O., Duan, Y., Poulenard, A., Tagliasacchi, A., and Guibas,
  L.~J.
\newblock Vector neurons: A general framework for so (3)-equivariant networks.
\newblock In \emph{Proceedings of the IEEE/CVF International Conference on
  Computer Vision}, pp.\  12200--12209, 2021.

\bibitem[Deng et~al.(2009)Deng, Dong, Socher, Li, Li, and
  Fei-Fei]{imagenet_cvpr09}
Deng, J., Dong, W., Socher, R., Li, L.-J., Li, K., and Fei-Fei, L.
\newblock {ImageNet: A Large-Scale Hierarchical Image Database}.
\newblock In \emph{CVPR09}, 2009.

\bibitem[Dong et~al.(2020)Dong, Chen, Zhou, Hua, Zhang, and Yu]{dong2020self}
Dong, X., Chen, D., Zhou, H., Hua, G., Zhang, W., and Yu, N.
\newblock Self-robust 3d point recognition via gather-vector guidance.
\newblock In \emph{2020 IEEE/CVF Conference on Computer Vision and Pattern
  Recognition (CVPR)}, pp.\  11513--11521. IEEE, 2020.

\bibitem[Ghahremani et~al.(2020)Ghahremani, Tiddeman, Liu, and
  Behera]{ghahremani2020orderly}
Ghahremani, M., Tiddeman, B., Liu, Y., and Behera, A.
\newblock Orderly disorder in point cloud domain.
\newblock In \emph{{ECCV} {(28)}}, volume 12373 of \emph{Lecture Notes in
  Computer Science}, pp.\  494--509. Springer, 2020.

\bibitem[Goyal et~al.(2021)Goyal, Law, Liu, Newell, and
  Deng]{goyal2021simpleview}
Goyal, A., Law, H., Liu, B., Newell, A., and Deng, J.
\newblock Revisiting point cloud shape classification with a simple and
  effective baseline.
\newblock \emph{International Conference on Machine Learning}, 2021.

\bibitem[Guo et~al.(2020)Guo, Cai, Liu, Mu, Martin, and Hu]{guo2020pct}
Guo, M.-H., Cai, J.-X., Liu, Z.-N., Mu, T.-J., Martin, R.~R., and Hu, S.-M.
\newblock Pct: Point cloud transformer, 2020.

\bibitem[Hendrycks \& Dietterich(2019)Hendrycks and
  Dietterich]{hendrycks2019bench}
Hendrycks, D. and Dietterich, T.~G.
\newblock Benchmarking neural network robustness to common corruptions and
  perturbations.
\newblock In \emph{{ICLR} (Poster)}. OpenReview.net, 2019.

\bibitem[Hendrycks et~al.(2021{\natexlab{a}})Hendrycks, Basart, Mu, Kadavath,
  Wang, Dorundo, Desai, Zhu, Parajuli, Guo, Song, Steinhardt, and
  Gilmer]{hendrycks2021many}
Hendrycks, D., Basart, S., Mu, N., Kadavath, S., Wang, F., Dorundo, E., Desai,
  R., Zhu, T., Parajuli, S., Guo, M., Song, D., Steinhardt, J., and Gilmer, J.
\newblock The many faces of robustness: A critical analysis of
  out-of-distribution generalization.
\newblock \emph{ICCV}, 2021{\natexlab{a}}.

\bibitem[Hendrycks et~al.(2021{\natexlab{b}})Hendrycks, Zhao, Basart,
  Steinhardt, and Song]{hendrycks2021nae}
Hendrycks, D., Zhao, K., Basart, S., Steinhardt, J., and Song, D.
\newblock Natural adversarial examples.
\newblock \emph{CVPR}, 2021{\natexlab{b}}.

\bibitem[Kim et~al.(2021)Kim, Lee, Hwang, Lee, Hwang, and
  Kim]{kim2021pointwolf}
Kim, S., Lee, S., Hwang, D., Lee, J., Hwang, S.~J., and Kim, H.~J.
\newblock Point cloud augmentation with weighted local transformations.
\newblock In \emph{Proceedings of the IEEE/CVF International Conference on
  Computer Vision (ICCV)}, pp.\  548--557, October 2021.

\bibitem[Lee et~al.(2021)Lee, Lee, Lee, Lee, Lee, Woo, and Lee]{lee2021rsmix}
Lee, D., Lee, J., Lee, J., Lee, H., Lee, M., Woo, S., and Lee, S.
\newblock Regularization strategy for point cloud via rigidly mixed sample.
\newblock In \emph{{CVPR}}, pp.\  15900--15909. Computer Vision Foundation /
  {IEEE}, 2021.

\bibitem[Li et~al.(2020)Li, Li, Heng, and Fu]{li2020pointaugment}
Li, R., Li, X., Heng, P., and Fu, C.
\newblock Pointaugment: An auto-augmentation framework for point cloud
  classification.
\newblock In \emph{{CVPR}}, pp.\  6377--6386. Computer Vision Foundation /
  {IEEE}, 2020.

\bibitem[Liu et~al.(2021)Liu, Jia, and Gong]{liu2021pointguard}
Liu, H., Jia, J., and Gong, N.~Z.
\newblock Pointguard: Provably robust 3d point cloud classification.
\newblock In \emph{Proceedings of the IEEE/CVF Conference on Computer Vision
  and Pattern Recognition}, pp.\  6186--6195, 2021.

\bibitem[Liu et~al.(2019)Liu, Fan, Xiang, and Pan]{liu2019rscnn}
Liu, Y., Fan, B., Xiang, S., and Pan, C.
\newblock Relation-shape convolutional neural network for point cloud analysis.
\newblock In \emph{IEEE Conference on Computer Vision and Pattern Recognition
  (CVPR)}, pp.\  8895--8904, 2019.

\bibitem[Mazur \& Lempitsky(2021)Mazur and
  Lempitsky]{mazur2021cloudtransformers}
Mazur, K. and Lempitsky, V.
\newblock Cloud transformers: A universal approach to point cloud processing
  tasks.
\newblock In \emph{International Conference on Computer Vision (ICCV)}, 2021.

\bibitem[Ortega et~al.(2018)Ortega, Frossard, Kova{\v{c}}evi{\'c}, Moura, and
  Vandergheynst]{ortega2018graph}
Ortega, A., Frossard, P., Kova{\v{c}}evi{\'c}, J., Moura, J.~M., and
  Vandergheynst, P.
\newblock Graph signal processing: Overview, challenges, and applications.
\newblock \emph{Proceedings of the IEEE}, 106\penalty0 (5):\penalty0 808--828,
  2018.

\bibitem[Qi et~al.(2017{\natexlab{a}})Qi, Yi, Su, and
  Guibas]{qi2017pointnetplusplus}
Qi, C., Yi, L., Su, H., and Guibas, L.~J.
\newblock Pointnet++: Deep hierarchical feature learning on point sets in a
  metric space.
\newblock In \emph{NIPS}, 2017{\natexlab{a}}.

\bibitem[Qi et~al.(2016)Qi, Su, Nie{\ss}ner, Dai, Yan, and
  Guibas]{qi2016volumetric}
Qi, C.~R., Su, H., Nie{\ss}ner, M., Dai, A., Yan, M., and Guibas, L.~J.
\newblock Volumetric and multi-view cnns for object classification on 3d data.
\newblock In \emph{Proceedings of the IEEE conference on computer vision and
  pattern recognition}, pp.\  5648--5656, 2016.

\bibitem[Qi et~al.(2017{\natexlab{b}})Qi, Su, Mo, and Guibas]{qi2017pointnet}
Qi, C.~R., Su, H., Mo, K., and Guibas, L.~J.
\newblock Pointnet: Deep learning on point sets for 3d classification and
  segmentation.
\newblock In \emph{Proceedings of the IEEE conference on computer vision and
  pattern recognition}, pp.\  652--660, 2017{\natexlab{b}}.

\bibitem[Recht et~al.(2019)Recht, Roelofs, Schmidt, and Shankar]{recht2019do}
Recht, B., Roelofs, R., Schmidt, L., and Shankar, V.
\newblock Do imagenet classifiers generalize to imagenet?
\newblock In \emph{{ICML}}, volume~97 of \emph{Proceedings of Machine Learning
  Research}, pp.\  5389--5400. {PMLR}, 2019.

\bibitem[Reizenstein et~al.(2021)Reizenstein, Shapovalov, Henzler, Sbordone,
  Labatut, and Novotny]{reizenstein21co3d}
Reizenstein, J., Shapovalov, R., Henzler, P., Sbordone, L., Labatut, P., and
  Novotny, D.
\newblock Common objects in 3d: Large-scale learning and evaluation of
  real-life 3d category reconstruction.
\newblock In \emph{International Conference on Computer Vision}, 2021.

\bibitem[Sandryhaila \& Moura(2014)Sandryhaila and
  Moura]{sandryhaila2014discrete}
Sandryhaila, A. and Moura, J.~M.
\newblock Discrete signal processing on graphs: Frequency analysis.
\newblock \emph{IEEE Transactions on Signal Processing}, 62\penalty0
  (12):\penalty0 3042--3054, 2014.

\bibitem[Simonovsky \& Komodakis(2017)Simonovsky and
  Komodakis]{simonovsky2017ecc}
Simonovsky, M. and Komodakis, N.
\newblock Dynamic edge-conditioned filters in convolutional neural networks on
  graphs.
\newblock In \emph{Proceedings of the IEEE Conference on Computer Vision and
  Pattern Recognition (CVPR)}, July 2017.

\bibitem[Sun et~al.(2021)Sun, Cao, Choy, Yu, Anandkumar, Mao, and
  Xiao]{sun2021adversarially}
Sun, J., Cao, Y., Choy, C., Yu, Z., Anandkumar, A., Mao, Z.~M., and Xiao, C.
\newblock Adversarially robust 3d point cloud recognition using
  self-supervisions.
\newblock In \emph{Thirty-Fifth Conference on Neural Information Processing
  Systems}, 2021.

\bibitem[Taghanaki et~al.(2020)Taghanaki, Luo, Zhang, Wang, Jayaraman, and
  Jatavallabhula]{taghanaki2020robustpointset}
Taghanaki, S.~A., Luo, J., Zhang, R., Wang, Y., Jayaraman, P.~K., and
  Jatavallabhula, K.~M.
\newblock Robustpointset: A dataset for benchmarking robustness of point cloud
  classifiers.
\newblock \emph{arXiv preprint arXiv:2011.11572}, 2020.

\bibitem[Uy et~al.(2019)Uy, Pham, Hua, Nguyen, and Yeung]{uy2019scanobjectnn}
Uy, M.~A., Pham, Q., Hua, B., Nguyen, D.~T., and Yeung, S.
\newblock Revisiting point cloud classification: {A} new benchmark dataset and
  classification model on real-world data.
\newblock In \emph{{ICCV}}, pp.\  1588--1597. {IEEE}, 2019.

\bibitem[Wang et~al.(2021)Wang, Liu, Yue, Lasenby, and Kusner]{wang2021occo}
Wang, H., Liu, Q., Yue, X., Lasenby, J., and Kusner, M.~J.
\newblock Unsupervised point cloud pre-training via occlusion completion.
\newblock In \emph{International Conference on Computer Vision, ICCV}, 2021.

\bibitem[Wang et~al.(2019)Wang, Sun, Liu, Sarma, Bronstein, and
  Solomon]{wang2019dgcnn}
Wang, Y., Sun, Y., Liu, Z., Sarma, S.~E., Bronstein, M.~M., and Solomon, J.~M.
\newblock Dynamic graph cnn for learning on point clouds.
\newblock \emph{ACM Transactions on Graphics (TOG)}, 2019.

\bibitem[Wu et~al.(2019)Wu, Zhou, Zhao, Yue, and Keutzer]{wu2019squeezesegv2}
Wu, B., Zhou, X., Zhao, S., Yue, X., and Keutzer, K.
\newblock Squeezesegv2: Improved model structure and unsupervised domain
  adaptation for road-object segmentation from a lidar point cloud.
\newblock In \emph{2019 International Conference on Robotics and Automation
  (ICRA)}, pp.\  4376--4382. IEEE, 2019.

\bibitem[Wu et~al.(2015)Wu, Song, Khosla, Yu, Zhang, Tang, and Xiao]{wu20153d}
Wu, Z., Song, S., Khosla, A., Yu, F., Zhang, L., Tang, X., and Xiao, J.
\newblock 3d shapenets: A deep representation for volumetric shapes.
\newblock In \emph{Proceedings of the IEEE conference on computer vision and
  pattern recognition}, pp.\  1912--1920, 2015.

\bibitem[Xiang et~al.(2021)Xiang, Zhang, Song, Yu, and Cai]{xiang2021curvenet}
Xiang, T., Zhang, C., Song, Y., Yu, J., and Cai, W.
\newblock Walk in the cloud: Learning curves for point clouds shape analysis.
\newblock In \emph{Proceedings of the IEEE/CVF International Conference on
  Computer Vision (ICCV)}, pp.\  915--924, October 2021.

\bibitem[Xiao \& Wachs(2021)Xiao and Wachs]{xiao2021triangle}
Xiao, C. and Wachs, J.~P.
\newblock Triangle-net: Towards robustness in point cloud learning.
\newblock In \emph{{WACV}}, pp.\  826--835. {IEEE}, 2021.

\bibitem[Xu et~al.(2021{\natexlab{a}})Xu, Ding, Zhao, and Qi]{xu2021paconv}
Xu, M., Ding, R., Zhao, H., and Qi, X.
\newblock Paconv: Position adaptive convolution with dynamic kernel assembling
  on point clouds.
\newblock In \emph{CVPR}, 2021{\natexlab{a}}.

\bibitem[Xu et~al.(2021{\natexlab{b}})Xu, Zhang, Zhou, Xu, Qi, and
  Qiao]{xu2021gdanet}
Xu, M., Zhang, J., Zhou, Z., Xu, M., Qi, X., and Qiao, Y.
\newblock Learning geometry-disentangled representation for complementary
  understanding of 3d object point cloud.
\newblock In \emph{{AAAI}}, pp.\  3056--3064. {AAAI} Press, 2021{\natexlab{b}}.

\bibitem[Yan et~al.(2020)Yan, Zheng, Li, Wang, and Cui]{yan2020pointasnl}
Yan, X., Zheng, C., Li, Z., Wang, S., and Cui, S.
\newblock Pointasnl: Robust point clouds processing using nonlocal neural
  networks with adaptive sampling.
\newblock In \emph{Proceedings of the IEEE/CVF Conference on Computer Vision
  and Pattern Recognition}, pp.\  5589--5598, 2020.

\bibitem[Yin et~al.(2019)Yin, Lopes, Shlens, Cubuk, and Gilmer]{yin2019fourier}
Yin, D., Lopes, R.~G., Shlens, J., Cubuk, E.~D., and Gilmer, J.
\newblock A fourier perspective on model robustness in computer vision.
\newblock In \emph{NeurIPS}, pp.\  13255--13265, 2019.

\bibitem[Yu et~al.(2021)Yu, Tang, Rao, Huang, Zhou, and Lu]{yu2021pointbert}
Yu, X., Tang, L., Rao, Y., Huang, T., Zhou, J., and Lu, J.
\newblock Point-bert: Pre-training 3d point cloud transformers with masked
  point modeling.
\newblock \emph{arXiv preprint}, 2021.

\bibitem[Yun et~al.(2019)Yun, Han, Oh, Chun, Choe, and Yoo]{yun2019cutmix}
Yun, S., Han, D., Oh, S.~J., Chun, S., Choe, J., and Yoo, Y.
\newblock Cutmix: Regularization strategy to train strong classifiers with
  localizable features.
\newblock In \emph{Proceedings of the IEEE/CVF International Conference on
  Computer Vision}, pp.\  6023--6032, 2019.

\bibitem[Zhang et~al.(2018)Zhang, Ciss{\'{e}}, Dauphin, and
  Lopez{-}Paz]{zhang2018mixup}
Zhang, H., Ciss{\'{e}}, M., Dauphin, Y.~N., and Lopez{-}Paz, D.
\newblock mixup: Beyond empirical risk minimization.
\newblock In \emph{{ICLR} (Poster)}. OpenReview.net, 2018.

\bibitem[Zhang et~al.(2019)Zhang, Hua, Rosen, and Yeung]{Zhang2019RotationIC}
Zhang, Z., Hua, B.-S., Rosen, D.~W., and Yeung, S.-K.
\newblock Rotation invariant convolutions for 3d point clouds deep learning.
\newblock \emph{2019 International Conference on 3D Vision (3DV)}, pp.\
  204--213, 2019.

\bibitem[Zhao et~al.(2021)Zhao, Jiang, Jia, Torr, and Koltun]{zhao2021point}
Zhao, H., Jiang, L., Jia, J., Torr, P.~H., and Koltun, V.
\newblock Point transformer.
\newblock In \emph{Proceedings of the IEEE/CVF International Conference on
  Computer Vision}, pp.\  16259--16268, 2021.

\bibitem[Zhou et~al.(2019)Zhou, Chen, Zhang, Fang, Zhou, and Yu]{zhou2019dup}
Zhou, H., Chen, K., Zhang, W., Fang, H., Zhou, W., and Yu, N.
\newblock Dup-net: Denoiser and upsampler network for 3d adversarial point
  clouds defense.
\newblock In \emph{Proceedings of the IEEE/CVF International Conference on
  Computer Vision}, pp.\  1961--1970, 2019.

\end{thebibliography}
\bibliographystyle{icml2022}

\appendix
\onecolumn
\section{Corruptions and Severity Level Settings}
\begin{figure*}[t!]
\vskip 0.in
\begin{center}
\centerline{\includegraphics[width=\columnwidth]{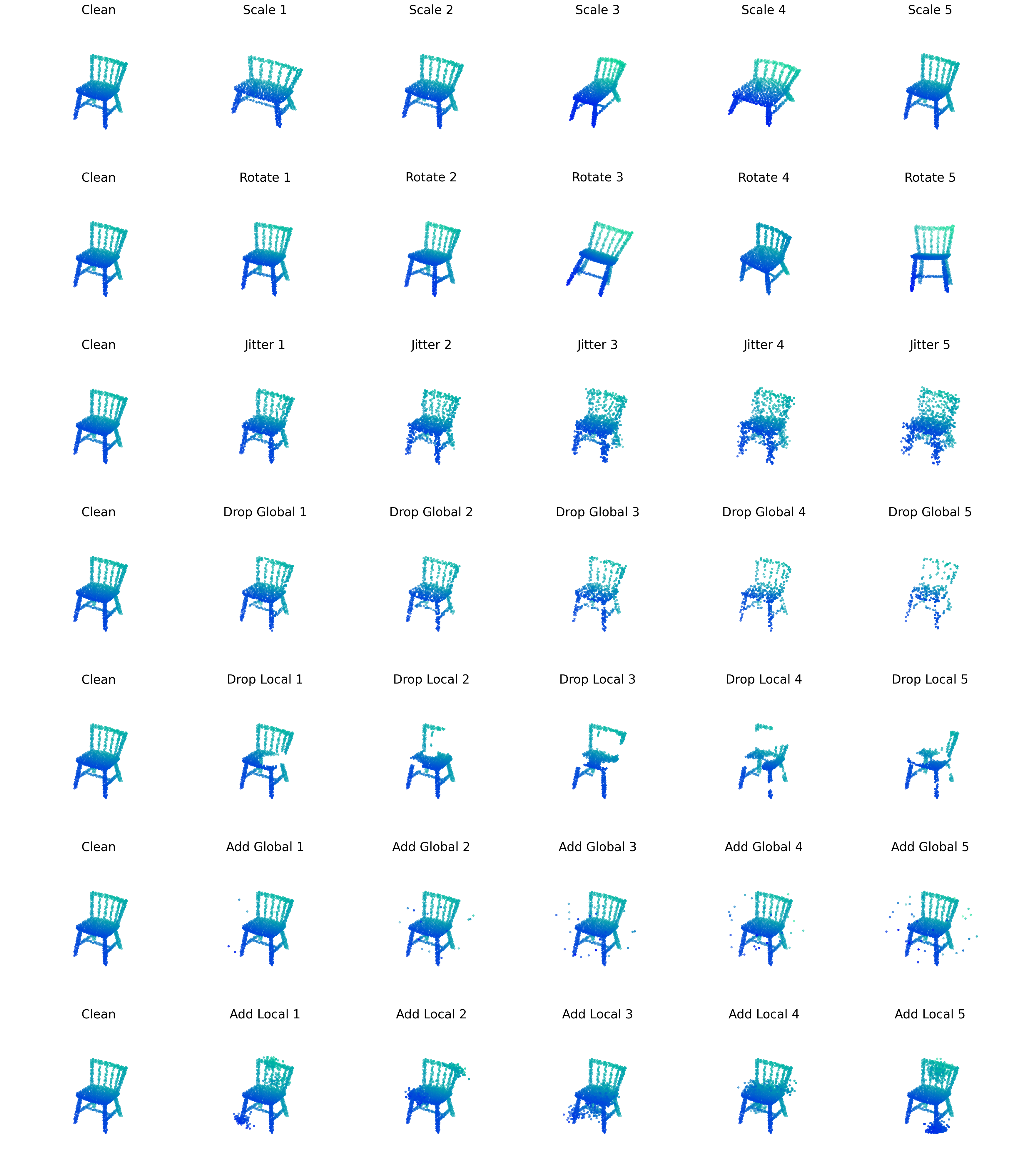}}
\caption{Corruptions on all levels. The severity of corruptions increases with the level. We average model's error on all levels for a comprehensive evaluation.}
\label{fig:difficulty}
\end{center}
\vskip -0.0in
\end{figure*}

We elaborate on the implementation of corruptions and severity level settings in this section. A visualization is shown in \autoref{fig:difficulty}.
\subsection{Jitter} We add a Gaussian noise $\epsilon \in \mathcal{N}(0, \sigma^2)$ to each of a point's X, Y, and Z coordinates, where $\sigma \in \{0.01, 0.02, 0.03, 0.04, 0.05\}$ for the five levels. 

\subsection{Scale} We apply random scaling to the X, Y, and Z axis respectively. The scaling coefficient for each axis are independently sampled as $s \sim \mathcal{U}(1/S, S)$, where $S \in \{1.6, 1.7, 1.8, 1.9, 2.0\}$ for the five levels. Point clouds are re-normalized to a unit sphere after scaling.

\subsection{Rotate} We randomly apply a rotation described by an X-Y-Z Euler angle $(\alpha, \beta, \gamma)$, where $\alpha, \beta, \gamma \sim \mathcal{U}(-\theta, \theta)$ and $\theta \in \{\pi/30, \pi/15, \pi/10, \pi/7.5, \pi/6\}$ for the five levels. Note that the sampling method does not guarantee a uniform SO(3) rotation sampling, but sufficient to cover a range of rotation variations.

\subsection{Drop Global} We randomly shuffle all points and drop the last $N*\rho$ points, where $N=1024$ is the number of points in the point cloud and $\rho \in \{0.25, 0.375, 0.5, 0.675, 0.75\}$ for all five levels.

\subsection{Drop Local} We drop $K$ points in total, where $K\in\{100, 200, 300, 400, 500\}$ for the five levels. We randomly choose $C$, the number of local parts to drop, by $C \in \mathcal{U}\{1, 8\}$. We further randomly assign $i$-th local part a cluster size $N_i$ so that $K=\sum_{i=1}^CN_i$. Then we repeat the following steps for $C$ times: we randomly select a point as the $i$-th local center, and drop its $N_i$-nearest neighbour points (including itself) from the point cloud.

\subsection{Add Global} We uniformly sample $K$ points inside a unit sphere and add them to the point cloud, where $K\in\{10, 20, 30, 40, 50\}$ for the five levels.

\subsection{Add Local} We add $K$ points in total, where $K\in\{100, 200, 300, 400, 500\}$ for the five levels.  We randomly shuffle the points, and select the first $C \in \mathcal{U}\{1, 8\}$  points as the local centers. We further randomly assign $i$-th local part a cluster size $N_i$ so that $K=\sum_{i=1}^CN_i$. Neighbouring point's X-Y-Z coordinates are generated from a Normal distribution $ \mathcal{N}(\boldsymbol{\mu}_i, \sigma_i^2\boldsymbol{\textrm{I}})$, where $\boldsymbol{\mu}_i$ is the i-th local center's X-Y-Z coordinate and $\sigma_i\in \mathcal{U}(0.075, 0.125)$. We then add each local part to the point cloud one by one.

\begin{figure}[t!]
\centering
\vskip 0.in
\begin{center}
\centerline{\includegraphics[width=\columnwidth]{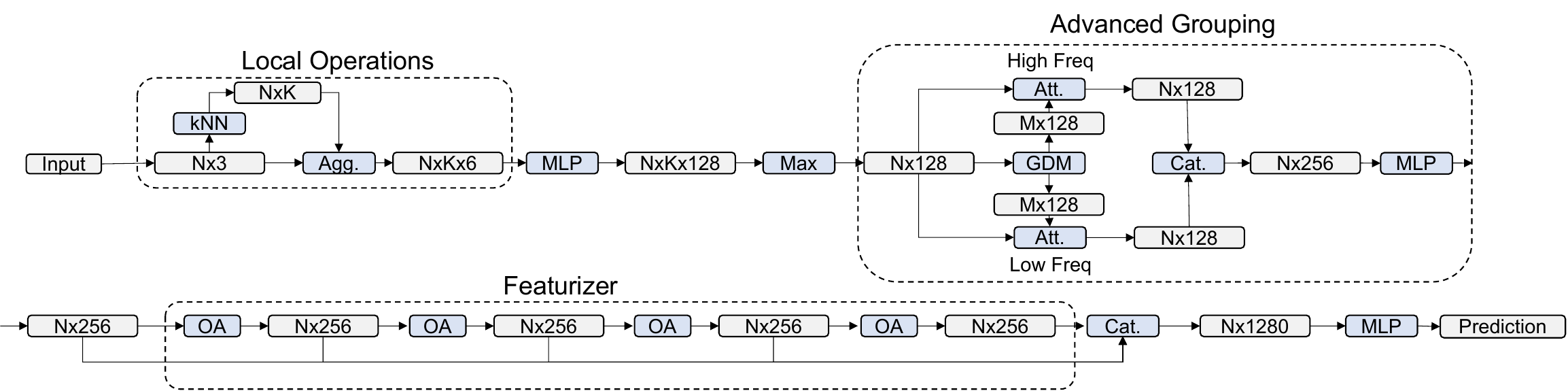}}
\vskip -0.0in
\caption{Detailed architecture of \model{}. We design \model{} following the conclusions we draw from the benchmark. It optimizes the use of existing building blocks in point cloud classifiers and serves as a strong baseline for corruption robustness.}
\label{fig:rpc}
\end{center}
\vskip -0.0in
\end{figure}

\section{Implementation Details}
We elaborate on implementation details for \model{} and \aug{} in this section.
\subsection{\model{}}
\subsubsection{Detailed Architecture}
We show a detailed architecture of \model{} in \autoref{fig:rpc}.
\subsubsection{Hyper-parameters}
For local operation, we use k=30 for the number of neighbors in k-NN. For, frequency grouping, we follow the default hyper-parameters in GDANet~\cite{xu2021gdanet}. The number of points in each frequency component is set to 256.
\subsubsection{Training}
We train the model for 250 epochs with a batch size of 32. We use SGD with momentum 0.9 for optimization. We use a cosine annealing scheduler to gradually decay the learning rate from 1e-2 to 1e-4. 

\subsection{\aug{}}
For the deformation step, we use the default hyper-parameters in PointWOLF~\cite{kim2021pointwolf}. We set the number of anchors to 4, sampling method to farthest point sampling, kernel bandwidth to 0.5, maximum local rotation range to 10 degrees, maximum local scaling to 3, and maximum local translation to 0.25. AugTune proposed along with PointWOLF is not used in training. For the mixing step, we use the default hyper-parameters in RSMix~\cite{lee2021rsmix}. We set RSMix probability to 0.5, $\beta$ to 1.0, and the maximum number of point modifications to 512. For training, the number of neighbors in k-NN is reduced to 20 and the number of epochs is increased to 500 for all methods.

\section{Additional Discussions}
\subsection{Correlation between \dataset{} mCE and ScanObjectNN OA}
\begin{figure}[!t]
\begin{center}
\centerline{\includegraphics[width=0.75\columnwidth]{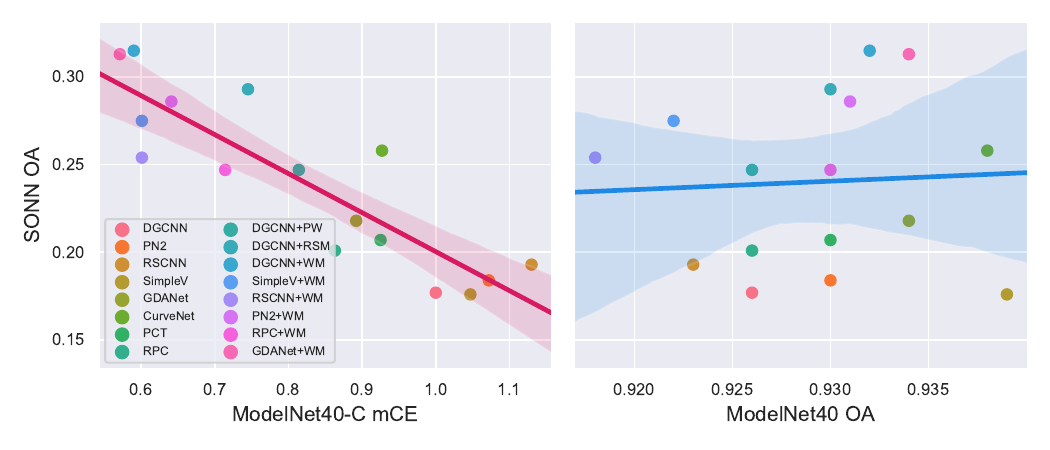}}
\caption{Correlation between \dataset{} mCE and ScanObjectNN OA.}
\label{fig:correlation}
\end{center}
\end{figure}

We additionally evaluate all models on ScanObjectNN~\cite{uy2019scanobjectnn}, and we use the \textit{OBJ\_BG PB\_T25} variant to include both background remains and bounding box inaccuracy. 
The results are shown in \autoref{fig:correlation}, and \dataset{} mCE strongly correlates to ScanObjectNN OA, while ModelNet40 OA has nearly no correlations.
Note that the results we report are lower than the results originally reported in ~\citet{uy2019scanobjectnn} due to different training protocols. \citet{uy2019scanobjectnn} uses random rotation and
per-point jitter in training while we follow the \textit{DGCNN protocol}~\cite{goyal2021simpleview}.

\begin{table}[ht]
\centering\small
\caption{Additional results of WolfMix.}
\label{tab:wolfmix_more}
\begin{tabular}{lc|c}
\toprule
{} & OA & mCE   \\
\midrule
PointNet++~\cite{qi2017pointnetplusplus} & 0.931 & 0.641 \\
RSCNN~\cite{liu2019rscnn} & 0.918 & 0.601 \\
SimpleView~\cite{goyal2021simpleview} & 0.922 & 0.676 \\
\bottomrule
\end{tabular}
\end{table}

\subsection{More results of Wolfmix}
We report additional results on RSCNN, SimpleView and PointNet++ with WolfMix in \autoref{tab:wolfmix_more}. The unequal benefits from augmentations motivate future research to explore the synergy between architecture and augmentation.

\subsection{Evaluation on specific techniques proposed for enhancing robustness}
There are a few methods for robustness enhancement. TriangleNet~\cite{xiao2021triangle}'s clean performance is not comparable to SoTA and PointASNL~\cite{yan2020pointasnl} requires a fixed number of points. Nonetheless, we manage to evaluate PointASNL with additional manual efforts and show the results in \autoref{tab:techniques}. PoinASNL shows outstanding performance to noisy jittering and achieves a competitive overall mCE result.

\begin{table}[!t]
\centering\small
\caption{More techniques.}
\label{tab:techniques}
\begin{tabular}{lc|c|ccccccc}
\toprule
{} &                               OA &                     \textbf{mCE} &                            Scale &                          Jitter &                          Drop-G &                           Drop-L &                           Add-G &                            Add-L &                           Rotate \\
\midrule
PointASNL~\cite{yan2020pointasnl} & 0.918 &  \cellcolor[HTML]{FFFFFF}0.959 &               \cellcolor[HTML]{FDE2E4}1.191 &               \cellcolor[HTML]{DFE7FD}0.687 &               \cellcolor[HTML]{FFFFFF}0.944 &               \cellcolor[HTML]{FFFFFF}0.826 &               \cellcolor[HTML]{FFFFFF}0.959 &               \cellcolor[HTML]{FFFFFF}0.953 &               \cellcolor[HTML]{FFFFFF}1.153 \\

Vector Neuron~\cite{deng2021vector} & 0.908 & \cellcolor[HTML]{FFFFFF}1.345 &               \cellcolor[HTML]{FFFFFF}1.287 &               \cellcolor[HTML]{FFFFFF}1.601 &               \cellcolor[HTML]{FDE2E4}1.875 &               \cellcolor[HTML]{FFFFFF}1.754 &               \cellcolor[HTML]{DFE7FD}0.902 &               \cellcolor[HTML]{FFFFFF}1.567 &  \cellcolor[HTML]{FFFFFF} \underline{0.428} \\

\bottomrule
\end{tabular}
\end{table}

\subsection{Evaluation on works designed for rotation robustness} Robustness to arbitrary SO(3) rotations is out of the scope of our benchmark where we examine common corruptions like small view angle variation. Nevertheless, we evaluate Vector Neurons~\cite{deng2021vector}, a rotation-invariant model, and show the results in \autoref{tab:techniques}. Vector Neuron achieves impressive robustness to rotational corruption, but under-performs to other types of corruption.

\begin{table*}
\centering\small
\caption{Full results for Overall Accuracy (OA). $\dagger$: randomly initialized. Bold: best in column. Underline: second best in column. Blue: best in row. Red: worst in row. mOA: average OA over all corruptions.}
\label{tab:benchmark_acc}
\begin{tabular}{lc|c|ccccccc}
\toprule
{} &                                       Clean $\uparrow$ &                                  mOA$\uparrow$ &                                       Scale &                                      Jitter &                                      Drop-G &                                      Drop-L &                                       Add-G &                                    Add-L &                                      Rotate \\
\midrule
DGCNN~\cite{wang2019dgcnn}                           &               \cellcolor[HTML]{FFFFFF}0.926 &            \cellcolor[HTML]{FFFFFF}0.764 &               \cellcolor[HTML]{DFE7FD}0.906 &               \cellcolor[HTML]{FDE2E4}0.684 &               \cellcolor[HTML]{FFFFFF}0.752 &               \cellcolor[HTML]{FFFFFF}0.793 &               \cellcolor[HTML]{FFFFFF}0.705 &            \cellcolor[HTML]{FFFFFF}0.725 &               \cellcolor[HTML]{FFFFFF}0.785 \\
PointNet~\cite{qi2017pointnet}                       &               \cellcolor[HTML]{FFFFFF}0.907 &            \cellcolor[HTML]{FFFFFF}0.658 &               \cellcolor[HTML]{DFE7FD}0.881 &               \cellcolor[HTML]{FFFFFF}0.797 &               \cellcolor[HTML]{FFFFFF}0.876 &               \cellcolor[HTML]{FFFFFF}0.778 &               \cellcolor[HTML]{FDE2E4}0.121 &            \cellcolor[HTML]{FFFFFF}0.562 &               \cellcolor[HTML]{FFFFFF}0.591 \\
PointNet++~\cite{qi2017pointnetplusplus}                     &               \cellcolor[HTML]{FFFFFF}0.930 &            \cellcolor[HTML]{FFFFFF}0.751 &               \cellcolor[HTML]{DFE7FD}0.918 &               \cellcolor[HTML]{FFFFFF}0.628 &               \cellcolor[HTML]{FFFFFF}0.841 &               \cellcolor[HTML]{FDE2E4}0.627 &               \cellcolor[HTML]{FFFFFF}0.819 &            \cellcolor[HTML]{FFFFFF}0.727 &               \cellcolor[HTML]{FFFFFF}0.698 \\
RSCNN~\cite{liu2019rscnn}                           &               \cellcolor[HTML]{FFFFFF}0.923 &            \cellcolor[HTML]{FFFFFF}0.739 &               \cellcolor[HTML]{DFE7FD}0.899 &               \cellcolor[HTML]{FDE2E4}0.630 &               \cellcolor[HTML]{FFFFFF}0.800 &               \cellcolor[HTML]{FFFFFF}0.686 &               \cellcolor[HTML]{FFFFFF}0.790 &            \cellcolor[HTML]{FFFFFF}0.683 &               \cellcolor[HTML]{FFFFFF}0.682 \\
SimpleView~\cite{goyal2021simpleview}                      &     \cellcolor[HTML]{FFFFFF}{0.939} &            \cellcolor[HTML]{FFFFFF}0.757 &               \cellcolor[HTML]{DFE7FD}0.918 &               \cellcolor[HTML]{FFFFFF}0.774 &               \cellcolor[HTML]{FDE2E4}0.692 &               \cellcolor[HTML]{FFFFFF}0.719 &               \cellcolor[HTML]{FFFFFF}0.710 &            \cellcolor[HTML]{FFFFFF}0.768 &               \cellcolor[HTML]{FFFFFF}0.717 \\
GDANet~\cite{xu2021gdanet}                          &               \cellcolor[HTML]{FFFFFF}0.934 &            \cellcolor[HTML]{FFFFFF}0.789 &     \cellcolor[HTML]{DFE7FD}{0.922} &               \cellcolor[HTML]{FDE2E4}0.735 &               \cellcolor[HTML]{FFFFFF}0.803 &               \cellcolor[HTML]{FFFFFF}0.815 &               \cellcolor[HTML]{FFFFFF}0.743 &            \cellcolor[HTML]{FFFFFF}0.715 &               \cellcolor[HTML]{FFFFFF}0.789 \\
CurveNet~\cite{xiang2021curvenet}                        &  \cellcolor[HTML]{FFFFFF}{0.938} &            \cellcolor[HTML]{FFFFFF}0.779 &               \cellcolor[HTML]{DFE7FD}0.918 &               \cellcolor[HTML]{FFFFFF}0.771 &               \cellcolor[HTML]{FFFFFF}0.824 &               \cellcolor[HTML]{FFFFFF}0.788 &               \cellcolor[HTML]{FDE2E4}0.603 &            \cellcolor[HTML]{FFFFFF}0.725 &               \cellcolor[HTML]{FFFFFF}0.826 \\
PAConv~\cite{xu2021paconv}                          &               \cellcolor[HTML]{FFFFFF}0.936 &            \cellcolor[HTML]{FFFFFF}0.730 &               \cellcolor[HTML]{DFE7FD}0.915 &               \cellcolor[HTML]{FDE2E4}0.537 &               \cellcolor[HTML]{FFFFFF}0.752 &               \cellcolor[HTML]{FFFFFF}0.792 &               \cellcolor[HTML]{FFFFFF}0.680 &            \cellcolor[HTML]{FFFFFF}0.643 &               \cellcolor[HTML]{FFFFFF}0.792 \\
PCT~\cite{guo2020pct}                             &               \cellcolor[HTML]{FFFFFF}0.930 &            \cellcolor[HTML]{FFFFFF}0.781 &               \cellcolor[HTML]{DFE7FD}0.918 &               \cellcolor[HTML]{FDE2E4}0.725 &               \cellcolor[HTML]{FFFFFF}0.869 &               \cellcolor[HTML]{FFFFFF}0.793 &               \cellcolor[HTML]{FFFFFF}0.770 &            \cellcolor[HTML]{FFFFFF}0.619 &               \cellcolor[HTML]{FFFFFF}0.776 \\
\model{} (Ours)                       &               \cellcolor[HTML]{FFFFFF}0.930 &            \cellcolor[HTML]{FFFFFF}0.795 &  \cellcolor[HTML]{DFE7FD}{0.921} &               \cellcolor[HTML]{FDE2E4}0.718 &               \cellcolor[HTML]{FFFFFF}0.878 &               \cellcolor[HTML]{FFFFFF}0.835 &               \cellcolor[HTML]{FFFFFF}0.726 &            \cellcolor[HTML]{FFFFFF}0.722 &               \cellcolor[HTML]{FFFFFF}0.768 \\
\midrule 
DGCNN+OcCo~\cite{wang2021occo}                      &               \cellcolor[HTML]{FFFFFF}0.922 &            \cellcolor[HTML]{FFFFFF}0.766 &               \cellcolor[HTML]{DFE7FD}0.849 &               \cellcolor[HTML]{FFFFFF}0.794 &               \cellcolor[HTML]{FFFFFF}0.776 &               \cellcolor[HTML]{FFFFFF}0.785 &               \cellcolor[HTML]{FDE2E4}0.574 &            \cellcolor[HTML]{FFFFFF}0.767 &               \cellcolor[HTML]{FFFFFF}0.820 \\
Point-BERT\textsuperscript{$\dagger$}                     &               \cellcolor[HTML]{FFFFFF}0.919 &            \cellcolor[HTML]{FFFFFF}0.678 &               \cellcolor[HTML]{DFE7FD}0.912 &               \cellcolor[HTML]{FFFFFF}0.688 &               \cellcolor[HTML]{FFFFFF}0.777 &               \cellcolor[HTML]{FFFFFF}0.732 &               \cellcolor[HTML]{FDE2E4}0.311 &            \cellcolor[HTML]{FFFFFF}0.626 &               \cellcolor[HTML]{FFFFFF}0.697 \\
Point-BERT~\cite{yu2021pointbert}                      &               \cellcolor[HTML]{FFFFFF}0.922 &            \cellcolor[HTML]{FFFFFF}0.693 &               \cellcolor[HTML]{DFE7FD}0.912 &               \cellcolor[HTML]{FFFFFF}0.602 &               \cellcolor[HTML]{FFFFFF}0.829 &               \cellcolor[HTML]{FFFFFF}0.762 &               \cellcolor[HTML]{FDE2E4}0.430 &            \cellcolor[HTML]{FFFFFF}0.604 &               \cellcolor[HTML]{FFFFFF}0.715 \\
\midrule PN2+PointMixUp~\cite{chen2020pointmixup}                     &               \cellcolor[HTML]{FFFFFF}0.915 &            \cellcolor[HTML]{FFFFFF}0.785 &               \cellcolor[HTML]{FFFFFF}0.843 &               \cellcolor[HTML]{FFFFFF}0.775 &               \cellcolor[HTML]{FFFFFF}0.801 &               \cellcolor[HTML]{FDE2E4}0.625 &               \cellcolor[HTML]{DFE7FD}0.865 &            \cellcolor[HTML]{FFFFFF}0.831 &               \cellcolor[HTML]{FFFFFF}0.757 \\
DGCNN+PW~\cite{kim2021pointwolf}                 &               \cellcolor[HTML]{FFFFFF}0.926 &            \cellcolor[HTML]{FFFFFF}0.809 &               \cellcolor[HTML]{DFE7FD}0.913 &               \cellcolor[HTML]{FDE2E4}0.727 &               \cellcolor[HTML]{FFFFFF}0.755 &               \cellcolor[HTML]{FFFFFF}0.819 &               \cellcolor[HTML]{FFFFFF}0.762 &            \cellcolor[HTML]{FFFFFF}0.790 &               \cellcolor[HTML]{FFFFFF}0.897 \\
DGCNN+RSMix~\cite{lee2021rsmix}                     &               \cellcolor[HTML]{FFFFFF}0.930 &            \cellcolor[HTML]{FFFFFF}0.839 &               \cellcolor[HTML]{FFFFFF}0.876 &               \cellcolor[HTML]{FDE2E4}0.724 &               \cellcolor[HTML]{FFFFFF}0.838 &               \cellcolor[HTML]{FFFFFF}0.878 &     \cellcolor[HTML]{DFE7FD}{0.917} &            \cellcolor[HTML]{FFFFFF}0.827 &               \cellcolor[HTML]{FFFFFF}0.813 \\
DGCNN+\aug{} (Ours)                   &               \cellcolor[HTML]{FFFFFF}0.932 &  \cellcolor[HTML]{FFFFFF}{0.871} &               \cellcolor[HTML]{FFFFFF}0.907 &               \cellcolor[HTML]{FDE2E4}0.774 &               \cellcolor[HTML]{FFFFFF}0.827 &               \cellcolor[HTML]{FFFFFF}0.881 &  \cellcolor[HTML]{DFE7FD}{0.916} &  \cellcolor[HTML]{FFFFFF}{0.886} &  \cellcolor[HTML]{FFFFFF}{0.903} \\
\midrule
PointNet+\aug{}                &               \cellcolor[HTML]{FFFFFF}0.884 &            \cellcolor[HTML]{FFFFFF}0.743 &               \cellcolor[HTML]{FFFFFF}0.801 &  \cellcolor[HTML]{FFFFFF}{0.850} &               \cellcolor[HTML]{DFE7FD}0.857 &               \cellcolor[HTML]{FFFFFF}0.776 &               \cellcolor[HTML]{FDE2E4}0.343 &            \cellcolor[HTML]{FFFFFF}0.807 &               \cellcolor[HTML]{FFFFFF}0.768 \\
PCT+\aug{}                     &               \cellcolor[HTML]{FFFFFF}0.934 &            \cellcolor[HTML]{FFFFFF}0.873 &               \cellcolor[HTML]{FFFFFF}0.906 &               \cellcolor[HTML]{FDE2E4}0.730 &  \cellcolor[HTML]{FFFFFF}{0.906} &               \cellcolor[HTML]{FFFFFF}0.898 &               \cellcolor[HTML]{DFE7FD}0.912 &            \cellcolor[HTML]{FFFFFF}0.861 &  \cellcolor[HTML]{FFFFFF}{0.895} \\
GDANet+\aug{}                     &               \cellcolor[HTML]{FFFFFF}0.934 &  \cellcolor[HTML]{FFFFFF}\textrm{0.871} &               \cellcolor[HTML]{DFE7FD}0.915 &               \cellcolor[HTML]{FDE2E4}0.721 &               \cellcolor[HTML]{FFFFFF}0.868 &     \cellcolor[HTML]{FFFFFF}{0.886} &               \cellcolor[HTML]{FFFFFF}0.910 &  \cellcolor[HTML]{FFFFFF}{0.886} &     \cellcolor[HTML]{FFFFFF}{0.912} \\
\model{}+\aug{}               &               \cellcolor[HTML]{FFFFFF}0.933 &            \cellcolor[HTML]{FFFFFF}0.865 &               \cellcolor[HTML]{DFE7FD}0.905 &               \cellcolor[HTML]{FDE2E4}0.694 &               \cellcolor[HTML]{FFFFFF}0.895 &               \cellcolor[HTML]{FFFFFF}0.894 &               \cellcolor[HTML]{FFFFFF}0.902 &            \cellcolor[HTML]{FFFFFF}0.868 &               \cellcolor[HTML]{FFFFFF}0.897 \\
\bottomrule
\end{tabular}
\end{table*}

\begin{table*}
\centering\small
\caption{Full results for Relative mCE. $\dagger$: random initialized. Bold: best in column. Underline: second best in column. Blue: best in row. Red: worst in row.}
\label{tab:benchmark_rmce}
\begin{tabular}{l|c|ccccccc}
\toprule
{} &                                     RmCE $\downarrow$ &                                       Scale &                                      Jitter &                                      Drop-G &                                      Drop-L &                                       Add-G &                                       Add-L &                                      Rotate \\
\midrule
DGCNN~\cite{wang2019dgcnn}                           &               \cellcolor[HTML]{FFFFFF}1.000 &               \cellcolor[HTML]{DFE7FD}1.000 &               \cellcolor[HTML]{DFE7FD}1.000 &               \cellcolor[HTML]{DFE7FD}1.000 &               \cellcolor[HTML]{DFE7FD}1.000 &               \cellcolor[HTML]{DFE7FD}1.000 &               \cellcolor[HTML]{DFE7FD}1.000 &               \cellcolor[HTML]{DFE7FD}1.000 \\
PointNet~\cite{qi2017pointnet}                        &               \cellcolor[HTML]{FFFFFF}1.488 &               \cellcolor[HTML]{FFFFFF}1.300 &               \cellcolor[HTML]{FFFFFF}0.455 &               \cellcolor[HTML]{DFE7FD}0.178 &               \cellcolor[HTML]{FFFFFF}0.970 &               \cellcolor[HTML]{FDE2E4}3.557 &               \cellcolor[HTML]{FFFFFF}1.716 &               \cellcolor[HTML]{FFFFFF}2.241 \\
PointNet++~\cite{qi2017pointnetplusplus}                      &               \cellcolor[HTML]{FFFFFF}1.114 &               \cellcolor[HTML]{FFFFFF}0.600 &               \cellcolor[HTML]{FFFFFF}1.248 &               \cellcolor[HTML]{FFFFFF}0.511 &               \cellcolor[HTML]{FDE2E4}2.278 &               \cellcolor[HTML]{DFE7FD}0.502 &               \cellcolor[HTML]{FFFFFF}1.010 &               \cellcolor[HTML]{FFFFFF}1.645 \\
RSCNN~\cite{liu2019rscnn}                           &               \cellcolor[HTML]{FFFFFF}1.201 &               \cellcolor[HTML]{FFFFFF}1.200 &               \cellcolor[HTML]{FFFFFF}1.211 &               \cellcolor[HTML]{FFFFFF}0.707 &               \cellcolor[HTML]{FDE2E4}1.782 &               \cellcolor[HTML]{DFE7FD}0.602 &               \cellcolor[HTML]{FFFFFF}1.194 &               \cellcolor[HTML]{FFFFFF}1.709 \\
SimpleView~\cite{goyal2021simpleview}                      &               \cellcolor[HTML]{FFFFFF}1.181 &               \cellcolor[HTML]{FFFFFF}1.050 &               \cellcolor[HTML]{DFE7FD}0.682 &               \cellcolor[HTML]{FFFFFF}1.420 &               \cellcolor[HTML]{FDE2E4}1.654 &               \cellcolor[HTML]{FFFFFF}1.036 &               \cellcolor[HTML]{FFFFFF}0.851 &               \cellcolor[HTML]{FFFFFF}1.574 \\
GDANet~\cite{xu2021gdanet}                          &               \cellcolor[HTML]{FFFFFF}0.865 &               \cellcolor[HTML]{DFE7FD}0.600 &               \cellcolor[HTML]{FFFFFF}0.822 &               \cellcolor[HTML]{FFFFFF}0.753 &               \cellcolor[HTML]{FDE2E4}0.895 &               \cellcolor[HTML]{FFFFFF}0.864 &               \cellcolor[HTML]{FFFFFF}1.090 &               \cellcolor[HTML]{FFFFFF}1.028 \\
CurveNet~\cite{xiang2021curvenet}                        &               \cellcolor[HTML]{FFFFFF}0.978 &               \cellcolor[HTML]{FFFFFF}1.000 &               \cellcolor[HTML]{FFFFFF}0.690 &               \cellcolor[HTML]{DFE7FD}0.655 &               \cellcolor[HTML]{FFFFFF}1.128 &               \cellcolor[HTML]{FDE2E4}1.516 &               \cellcolor[HTML]{FFFFFF}1.060 &               \cellcolor[HTML]{FFFFFF}0.794 \\
PAConv~\cite{xu2021paconv}                          &               \cellcolor[HTML]{FFFFFF}1.211 &               \cellcolor[HTML]{DFE7FD}1.050 &               \cellcolor[HTML]{FDE2E4}1.649 &               \cellcolor[HTML]{FFFFFF}1.057 &               \cellcolor[HTML]{FFFFFF}1.083 &               \cellcolor[HTML]{FFFFFF}1.158 &               \cellcolor[HTML]{FFFFFF}1.458 &               \cellcolor[HTML]{FFFFFF}1.021 \\
PCT~\cite{guo2020pct}                             &               \cellcolor[HTML]{FFFFFF}0.884 &               \cellcolor[HTML]{FFFFFF}0.600 &               \cellcolor[HTML]{FFFFFF}0.847 &               \cellcolor[HTML]{DFE7FD}0.351 &               \cellcolor[HTML]{FDE2E4}1.030 &               \cellcolor[HTML]{FFFFFF}0.724 &               \cellcolor[HTML]{FFFFFF}1.547 &               \cellcolor[HTML]{FFFFFF}1.092 \\
\model{} (Ours)                       &               \cellcolor[HTML]{FFFFFF}0.778 &  \cellcolor[HTML]{FFFFFF}{0.450} &               \cellcolor[HTML]{FFFFFF}0.876 &               \cellcolor[HTML]{DFE7FD}0.299 &               \cellcolor[HTML]{FFFFFF}0.714 &               \cellcolor[HTML]{FDE2E4}0.923 &               \cellcolor[HTML]{FFFFFF}1.035 &               \cellcolor[HTML]{FFFFFF}1.149 \\
\midrule 
DGCNN+OcCo~\cite{wang2021occo}                      &               \cellcolor[HTML]{FFFFFF}1.302 &               \cellcolor[HTML]{FDE2E4}3.650 &               \cellcolor[HTML]{DFE7FD}0.529 &               \cellcolor[HTML]{FFFFFF}0.839 &               \cellcolor[HTML]{FFFFFF}1.030 &               \cellcolor[HTML]{FFFFFF}1.575 &               \cellcolor[HTML]{FFFFFF}0.771 &               \cellcolor[HTML]{FFFFFF}0.723 \\
Point-BERT\textsuperscript{$\dagger$}                  &               \cellcolor[HTML]{FFFFFF}1.330 &     \cellcolor[HTML]{DFE7FD}{0.350} &               \cellcolor[HTML]{FFFFFF}0.955 &               \cellcolor[HTML]{FFFFFF}0.816 &               \cellcolor[HTML]{FFFFFF}1.406 &               \cellcolor[HTML]{FDE2E4}2.751 &               \cellcolor[HTML]{FFFFFF}1.458 &               \cellcolor[HTML]{FFFFFF}1.574 \\
Point-BERT~\cite{yu2021pointbert}                      &               \cellcolor[HTML]{FFFFFF}1.262 &               \cellcolor[HTML]{DFE7FD}0.500 &               \cellcolor[HTML]{FFFFFF}1.322 &               \cellcolor[HTML]{FFFFFF}0.534 &               \cellcolor[HTML]{FFFFFF}1.203 &               \cellcolor[HTML]{FDE2E4}2.226 &               \cellcolor[HTML]{FFFFFF}1.582 &               \cellcolor[HTML]{FFFFFF}1.468 \\
\midrule PN2+PointMixUp~\cite{chen2020pointmixup}                 &               \cellcolor[HTML]{FFFFFF}1.254 &               \cellcolor[HTML]{FDE2E4}3.600 &               \cellcolor[HTML]{FFFFFF}0.579 &               \cellcolor[HTML]{FFFFFF}0.655 &               \cellcolor[HTML]{FFFFFF}2.180 &               \cellcolor[HTML]{DFE7FD}0.226 &               \cellcolor[HTML]{FFFFFF}0.418 &               \cellcolor[HTML]{FFFFFF}1.121 \\
DGCNN+PointWOLF~\cite{kim2021pointwolf}                 &               \cellcolor[HTML]{FFFFFF}0.698 &               \cellcolor[HTML]{DFE7FD}0.650 &               \cellcolor[HTML]{FFFFFF}0.822 &               \cellcolor[HTML]{FDE2E4}0.983 &               \cellcolor[HTML]{FFFFFF}0.805 &               \cellcolor[HTML]{FFFFFF}0.742 &               \cellcolor[HTML]{FFFFFF}0.677 &               \cellcolor[HTML]{FFFFFF}0.206 \\
DGCNN+RSMix~\cite{lee2021rsmix}                     &               \cellcolor[HTML]{FFFFFF}0.839 &               \cellcolor[HTML]{FDE2E4}2.700 &               \cellcolor[HTML]{FFFFFF}0.851 &               \cellcolor[HTML]{FFFFFF}0.529 &               \cellcolor[HTML]{FFFFFF}0.391 &     \cellcolor[HTML]{DFE7FD}{0.059} &               \cellcolor[HTML]{FFFFFF}0.512 &               \cellcolor[HTML]{FFFFFF}0.830 \\
DGCNN+\aug{} (Ours)                   &  \cellcolor[HTML]{FFFFFF}{0.485} &               \cellcolor[HTML]{FDE2E4}1.250 &               \cellcolor[HTML]{FFFFFF}0.653 &               \cellcolor[HTML]{FFFFFF}0.603 &               \cellcolor[HTML]{FFFFFF}0.383 &  \cellcolor[HTML]{DFE7FD}{0.072} &     \cellcolor[HTML]{FFFFFF}{0.229} &               \cellcolor[HTML]{FFFFFF}0.206 \\
\midrule
PCT+\aug{}                     &               \cellcolor[HTML]{FFFFFF}0.488 &               \cellcolor[HTML]{FDE2E4}1.400 &               \cellcolor[HTML]{FFFFFF}0.843 &               \cellcolor[HTML]{FFFFFF}0.161 &               \cellcolor[HTML]{FFFFFF}0.271 &               \cellcolor[HTML]{DFE7FD}0.100 &               \cellcolor[HTML]{FFFFFF}0.363 &               \cellcolor[HTML]{FFFFFF}0.277 \\
GDANet+\aug{}&     \cellcolor[HTML]{FFFFFF}{0.439} &               \cellcolor[HTML]{FDE2E4}0.950 &               \cellcolor[HTML]{FFFFFF}0.880 &               \cellcolor[HTML]{FFFFFF}0.379 &     \cellcolor[HTML]{FFFFFF}{0.361} &               \cellcolor[HTML]{DFE7FD}0.109 &  \cellcolor[HTML]{FFFFFF}{0.239} &  \cellcolor[HTML]{FFFFFF}{0.156} \\
\model{}+\aug{}               &               \cellcolor[HTML]{FFFFFF}0.517 &               \cellcolor[HTML]{FDE2E4}1.400 &               \cellcolor[HTML]{FFFFFF}0.988 &               \cellcolor[HTML]{FFFFFF}0.218 &               \cellcolor[HTML]{FFFFFF}0.293 &               \cellcolor[HTML]{DFE7FD}0.140 &               \cellcolor[HTML]{FFFFFF}0.323 &               \cellcolor[HTML]{FFFFFF}0.255 \\
\bottomrule
\end{tabular}
\end{table*}

\section{Full Results}
We show full results for the OA metric and the RmCE metric in \autoref{tab:benchmark_acc} and \autoref{tab:benchmark_rmce}.

\end{document}